\documentclass{article} 
\usepackage{behaviorworldgen,times}

\usepackage{amsmath,amsfonts,bm}

\def\eqref#1{equation~\ref{#1}}

\def\1{\bm{1}}

\DeclareMathAlphabet{\mathsfit}{\encodingdefault}{\sfdefault}{m}{sl}
\SetMathAlphabet{\mathsfit}{bold}{\encodingdefault}{\sfdefault}{bx}{n}

\usepackage{hyperref}
\usepackage{url}
\usepackage{booktabs}       
\usepackage{amsfonts}       
\usepackage{nicefrac}       
\usepackage{microtype}      
\usepackage{xcolor}         
\usepackage{float}
\usepackage{multirow}
\usepackage{adjustbox}
\usepackage{amsmath}
\usepackage{cleveref}
\usepackage{booktabs}
\usepackage{array}
\usepackage{graphicx}
\usepackage[table]{xcolor}
\usepackage{pifont}
\usepackage{caption}
\usepackage{placeins}
\usepackage{booktabs}
\usepackage[table]{xcolor}
\usepackage{caption}
\usepackage{pifont}
\usepackage{afterpage}
\usepackage{dsfont}

\newcolumntype{H}{@{}>{\setbox0=\hbox\bgroup}c<{\egroup}@{}}

\newenvironment{indented}
{\list{}{\setlength{\leftmargin}{2em}}\item\relax}
{\endlist}

\title{BehaviorWorldGen: Closing the Loop between Action Models and World Simulators via Controllable Behavior‑Aware Structured World Generation}

\author{\hyperlink{author-list}{AFARI World Model Team}}

\begin{document}

\maketitle

\begin{abstract}
Modern driving action models are increasingly improved in a self-improvement loop, where a learned world simulator imagines future observations and the resulting data is fed back to refine the action model. However, the bottleneck of this loop lies in the simulators' inability to generate behaviorally plausible responses by surrounding agents, making generated data both unrealistic in interaction and imbalanced in distribution. We introduce \textbf{BehaviorWorldGen}, a framework that closes the loop between action models and world simulators through controllable behavior-aware structured world generation. Its core component is \textbf{BehaviorFlow}, a meta-action-conditioned traffic-flow model that injects interpretable behavior controls and jointly generates multi-agent rollouts. BehaviorFlow realizes the specified agent behaviors while allowing surrounding vehicles to respond to the ego and to one another. The resulting rollouts are rendered by a world simulator into realistic multi-view observations, which are paired with corrected interaction-aware trajectories for action-model refinement. Since BehaviorWorldGen uses structured trajectories as the interface between its modules, it is compatible with diverse action models and world simulators. Experiments on world generation, scene extrapolation, and policy refinement demonstrate consistent improvements, with the largest benefits concentrated on difficult interactive scenarios.

\par\vspace{1em}
\small
\textbf{Date:} August 2026

\textbf{Project:}
\href{https://behaviorworldgen.github.io/}
{\texttt{https://behaviorworldgen.github.io/}}
\hspace{1.5em}
\end{abstract}

\section{Introduction}
End-to-end autonomous driving is increasingly built around \emph{action models}, such as end-to-end planners~\citep{uniad, jiang2023vad}, vision-language-action (VLA) models~\citep{tian2024drivevlm, zhou2026opendrivevla, wang2026chainflow}, and world-action models. These models map multimodal observations conditioned on navigation intent to future ego behavior. To move beyond passive data scaling, a growing line of work improves action models through a self-improvement loop with a learned world simulator~\citep{li2026world, guo2026vlaw, liu2026world}. The world simulator imagines future observations under a candidate ego behavior~\citep{gao2024vista, wang2024driving}. The action model then acts within these imagined futures, with the resulting data fed back for refinement through reinforcement learning or supervised fine-tuning~\citep{zhou2026autovla}. This loop promises stronger driving policies without relying solely on real-world collection.

In interactive scenarios such as cut-in, merging, yielding, or intersection negotiation, a good planning model depends on the ego trajectory geometry as well as the responses of surrounding agents. However, current world simulators are forward-dynamics models conditioned on the ego action alone. Generative world models synthesize future sensor data from maps, boxes, and the ego trajectory~\citep{gao2024magicdrive, wang2024drivedreamer, gao2024vista, Deng_2026_CVPR}. Reconstruction-based simulators instead replay a recorded scene with high-fidelity novel-view rendering~\citep{kerbl20233d, yan2024street, chen2025omnire, deng2026prosgnerf}. In both cases, surrounding agents are either replayed from history or produced as a by-product of generation. Their futures remain fixed with respect to the ego action, so the simulator cannot infer or control their behavior. Therefore, for interactive driving, the bottleneck lies in the simulator's inability to model behaviorally plausible responses from surrounding agents.

This limitation leads to two problems. First, surrounding agents do not truly react to the ego, which distorts the interaction feedback. A plan that would provoke a response in the real world can appear perfectly safe inside the simulator. Any action model refined on this feedback inherits the same distortion, whether through reinforcement learning or supervised fine-tuning. Second, the loop is undirected, so it mostly reproduces routine driving. It rarely surfaces the long-tail interactions that matter most, since these require coordinated multi-agent behaviors~\citep{ma2024unleashing}. The resulting data are unrealistic in interaction and imbalanced in distribution, which helps explain why rare interactive scenarios remain difficult to learn~\citep{ma2026correctad}.

To address this limitation, we introduce \textbf{BehaviorWorldGen}, a framework that couples action models with world simulators through controllable behavior-aware structured world generation. Its core component is \textbf{BehaviorFlow}, a meta-action-conditioned traffic-flow model that controls agent lifecycles together with high-level behaviors. BehaviorFlow attaches an interpretable frame-level meta-action to each agent, such as keeping lane, changing lane or turning~\citep{zhao2025autoregressive}. These meta-actions act as explicit behavioral handles that can be injected at will. Conditioned on the injected controls, BehaviorFlow realizes the specified behaviors in a multi-agent rollout where surrounding vehicles respond to the ego and to one another. Unlike controllable traffic models developed for offline simulation or benchmarking~\citep{zhong2023ctg, tan2023lctgen, philion2024trajeglish}, BehaviorFlow is designed as a bridge within the action-model self-improvement loop. Explicit behavior control allows rare interactions to be orchestrated on demand, with emphasis placed on scenarios where the current action model is weakest. This design turns a passive simulator into a source of targeted interactive training data.

BehaviorWorldGen closes the loop from the outside. An action model first predicts an ego trajectory, which is evaluated with scenario-level criteria to identify failure cases. For each failure case, the framework derives target meta-actions for BehaviorFlow. Conditioned on these controls, BehaviorFlow jointly evolves the ego vehicle alongside surrounding agents into a controllable rollout that captures the intended behaviors with their resulting interactions. A world simulator then renders the rollout into realistic multi-view observations. These observations are paired with the corrected interaction-aware trajectories before being added to the action model's training data. BehaviorWorldGen operates on structured trajectories at both interfaces. It accepts the output of any action model upstream, then passes its rollout to any world simulator downstream.

\noindent Our main contributions are as follows.
\begin{itemize}
	\item We identify a structural limitation in current self-improvement loops for interactive driving, where simulators cannot control surrounding-agent behaviors, resulting in unreliable interaction feedback with insufficient coverage of long-tail scenarios.
	\item We introduce \textbf{BehaviorWorldGen}, whose core \textbf{BehaviorFlow} model provides meta-action-conditioned control over agent lifecycles and high-level behaviors while connecting action models with world simulators in an outer loop.
	\item We show that \textbf{BehaviorWorldGen} improves diverse action models while remaining compatible with different world simulators, enabling scalable refinement on rare interactive scenarios.
\end{itemize}

\section{Related Works}
\subsection{World Simulator}
A world simulator provides the environment of the self-improvement loop by turning a driving scene and an ego action into future observations. One promising direction learns \emph{action-conditioned world models} that generate future sensor data under structured controls. DriveDreamer~\citep{wang2024drivedreamer} and MagicDrive~\citep{gao2024magicdrive} synthesize street-view videos from road maps, boxes, and camera poses. GAIA-1~\citep{hu2023gaia} and Vista~\citep{gao2024vista} scale generative driving world models with high fidelity and versatile action control. Drive-WM~\citep{wang2024driving} extends generation to multi-view futures and uses them to evaluate candidate maneuvers, and Delphi~\citep{ma2024unleashing} shows that controllable long video generation improves end-to-end planning. Closed-loop variants such as DriveArena~\citep{yang2025drivearena} and CausalDrive~\citep{yan2026causaldrive} render reactive observations as the policy acts. The second direction reconstructs the scene explicitly with 3D Gaussian Splatting~\citep{kerbl20233d}. Street Gaussians~\citep{yan2024street}, DrivingGaussian~\citep{zhou2024drivinggaussian}, and S3Gaussian~\citep{huang2024s3gaussian} decompose static backgrounds and dynamic objects for high-fidelity novel-view rendering. OmniRe~\citep{chen2025omnire} further models non-rigid actors through a Gaussian scene graph, and SplatAD~\citep{hess2025splatad} renders camera and LiDAR jointly in real time. World Engine~\citep{li2026world} reconstructs interactive environments from logs for post-training. 
\vspace{-0.1cm}

\subsection{Action Model}
We use \emph{action model} to denote any policy that maps observations and navigation intent to ego behavior. Early end-to-end planners unify perception, prediction, and planning in a single differentiable network. UniAD~\citep{uniad} organizes full-stack driving tasks around planning-oriented queries, and VAD~\citep{jiang2023vad} replaces dense rasterization with a vectorized scene representation for efficiency. Follow-ups such as VADv2~\citep{jiang2024vadv2}, SparseDrive~\citep{sun2025sparsedrive}, and Para-Drive~\citep{weng2024drive} refine this paradigm. GenAD~\citep{zheng2024genad} and DiffusionDrive~\citep{liao2025diffusiondrive} instead adopt generative models to capture multi-modal trajectory distributions. A second line couples driving with large vision-language models to form vision-language-action (VLA) models. DriveVLM~\citep{tian2024drivevlm} adds scene description and hierarchical reasoning. OpenDriveVLA~\citep{zhou2026opendrivevla} aligns visual tokens, ego states, and language commands into autoregressive action generation. AutoVLA~\citep{zhou2026autovla} explores adaptive reasoning and reinforcement fine-tuning, and ChainFlow-VLA~\citep{wang2026chainflow} decomposes planning into causal proposal generation and VLM-guided residual refinement. Recent world-action models further pair a policy with an action-conditioned world model and improve them together~\citep{liu2026world, guo2026vlaw}. 
\vspace{-0.1cm}

\subsection{Traffic Flow Model}
Traffic flow models generate the motion of multiple agents in a scene and form the interaction core of any simulator. Rule-based models such as IDM~\citep{treiber2000congested} are controllable but unrealistic, and cannot cover diverse interactive behaviors. Learning-based models improve realism by generating agent motion directly from data. TrafficSim~\citep{suo2021trafficsim} and TrafficGen~\citep{feng2023trafficgen} learn to initialize and roll out agents from logged scenarios. A controllable diffusion line then adds test-time control. CTG~\citep{zhong2023ctg} guides diffusion with signal-temporal-logic rules. CTG++~\citep{zhong2023ctgpp} replaces this guidance with LLM-generated cost functions. LCTGen~\citep{tan2023lctgen} and RealGen~\citep{ding2024realgen} generate scenarios from language or retrieved examples. A parallel line casts traffic modeling as next-token prediction. Trajeglish~\citep{philion2024trajeglish} and SMART~\citep{wu2024smart} tokenize agent motion and generate interactive rollouts autoregressively. MotionLM~\citep{seff2023motionlm} and InfGen~\citep{peng2025infgen} extend this to joint multi-agent prediction and long-horizon scenario generation. Most closely related, meta-action trajectory generation~\citep{zhao2025autoregressive} attaches a frame-level meta-action to each agent, which makes high-level behavior injectable while keeping motion physically feasible. 

\begin{figure}[t]
	\centering
	\includegraphics[width=1\hsize]{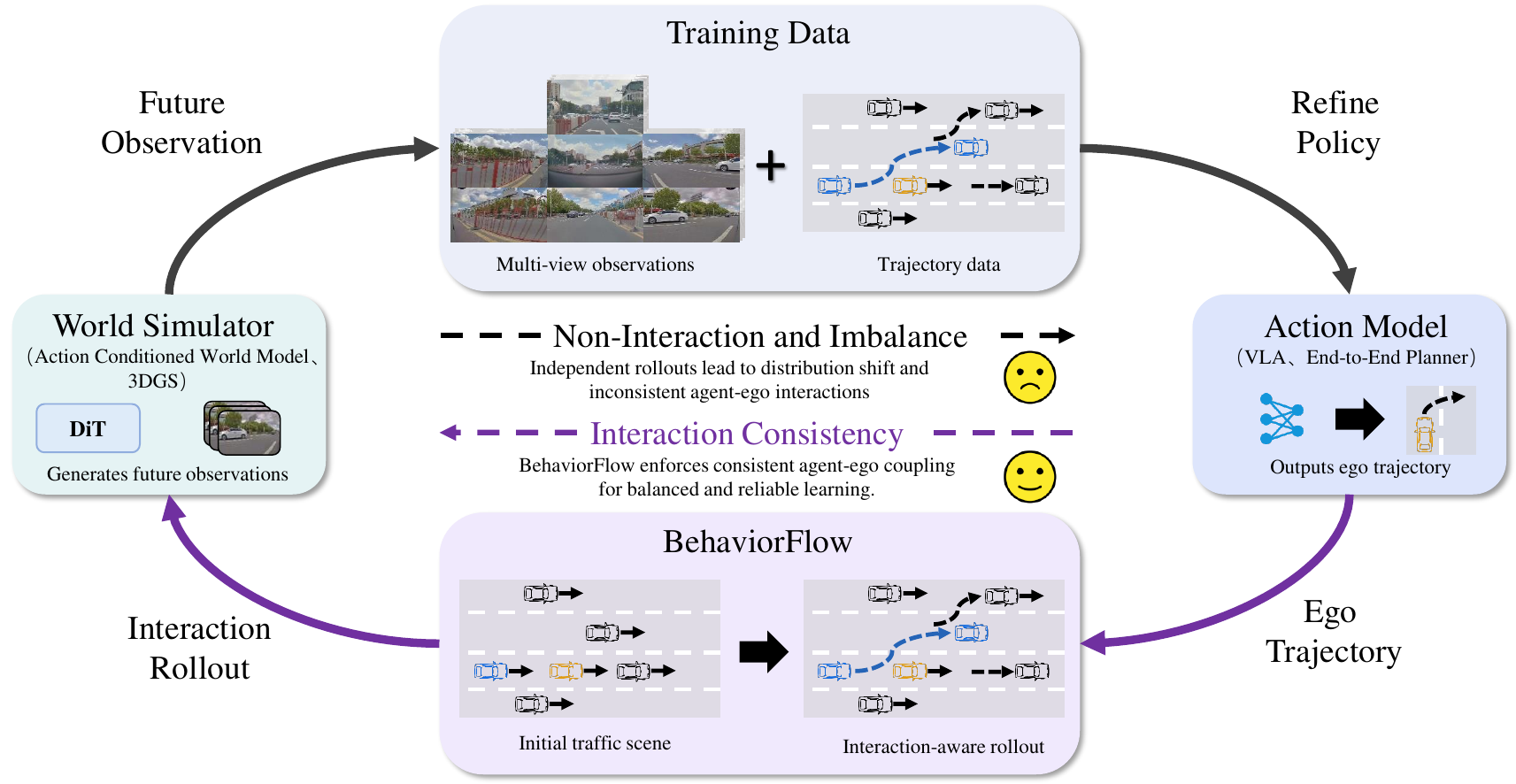}
	\caption{The overview of our framework.}
	\label{fig:overview}
    \vspace{-0.6cm}
\end{figure}

\section{Methodology}
\subsection{Overview}
\textbf{BehaviorWorldGen} improves the action model through a closed-loop data process that couples an action model, a \textbf{BehaviorFlow} traffic-flow model, and a world simulator. The key departure from prior self-improvement loops is that BehaviorFlow operates in an outer loop between the action model and the world simulator. Surrounding agents are guided by explicit meta-actions with interpretable semantics. As illustrated in Fig.~\ref{fig:overview}, the three modules exchange structured trajectories plus standard training data. This interface keeps each module agnostic to its counterpart.

We formalize the loop as follows. Given the current driving observation $\mathcal{O}$, the action model $\pi$ predicts an initial ego trajectory $\hat{Y}^{\mathrm{act}} = \pi(\mathcal{O})$. This trajectory is diagnosed against a set of scenario-level scores. Poor-performing scenarios are then converted into meta-action prompts before being passed to BehaviorFlow. Conditioned on the injected meta-actions, BehaviorFlow jointly evolves the ego vehicle with surrounding agents into an interaction-consistent rollout that realizes the specified behaviors while allowing surrounding vehicles to respond to the ego. The rollout is rasterized into a multi-view layout before the world simulator renders it as realistic multi-view RGB videos with synchronized LiDAR sequences. These observations are paired with corrected interaction-aware trajectories before being appended to the training data of the action model. This pairing transfers the correction back to the policy.

The three modules play complementary roles across the rest of this section. The world simulator (Sec.~\ref{sec:method_wm}) provides the environment of the loop. We instantiate it with either a generative action-conditioned world model or a reconstruction-based 3D Gaussian Splatting pipeline. The action model (Sec.~\ref{sec:method_am}) is the policy under improvement. It can be instantiated with end-to-end planners or vision-language-action (VLA) models. This setup demonstrates that the same loop can improve different policy methods. BehaviorFlow (Sec.~\ref{sec:behaviorflow}) is the core module for correcting interaction failures. It also rebalances the training distribution toward long-tail interactive scenarios that matter most for safety.

\subsection{World Simulator}\label{sec:method_wm}
The world simulator consumes a behavior-aware rollout and produces realistic observations for action-model training. We instantiate it with either an action-conditioned world model or a reconstruction-based 3D Gaussian Splatting pipeline. The action-conditioned world model (Sec.~\ref{sec:method_wm_arch}) generates future multi-view RGB videos together with LiDAR rangemaps from an HD-map layout. This design offers broad controllability with the ability to imagine scenes beyond recorded logs. The 3D Gaussian Splatting pipeline (Sec.~\ref{sec:method_wm_3dgs}) reconstructs the observed scene for high-fidelity novel-view rendering. We extend its limited reconstruction boundary using rollouts from BehaviorFlow.

\begin{figure}[t]
	\centering
	\includegraphics[width=1\hsize]{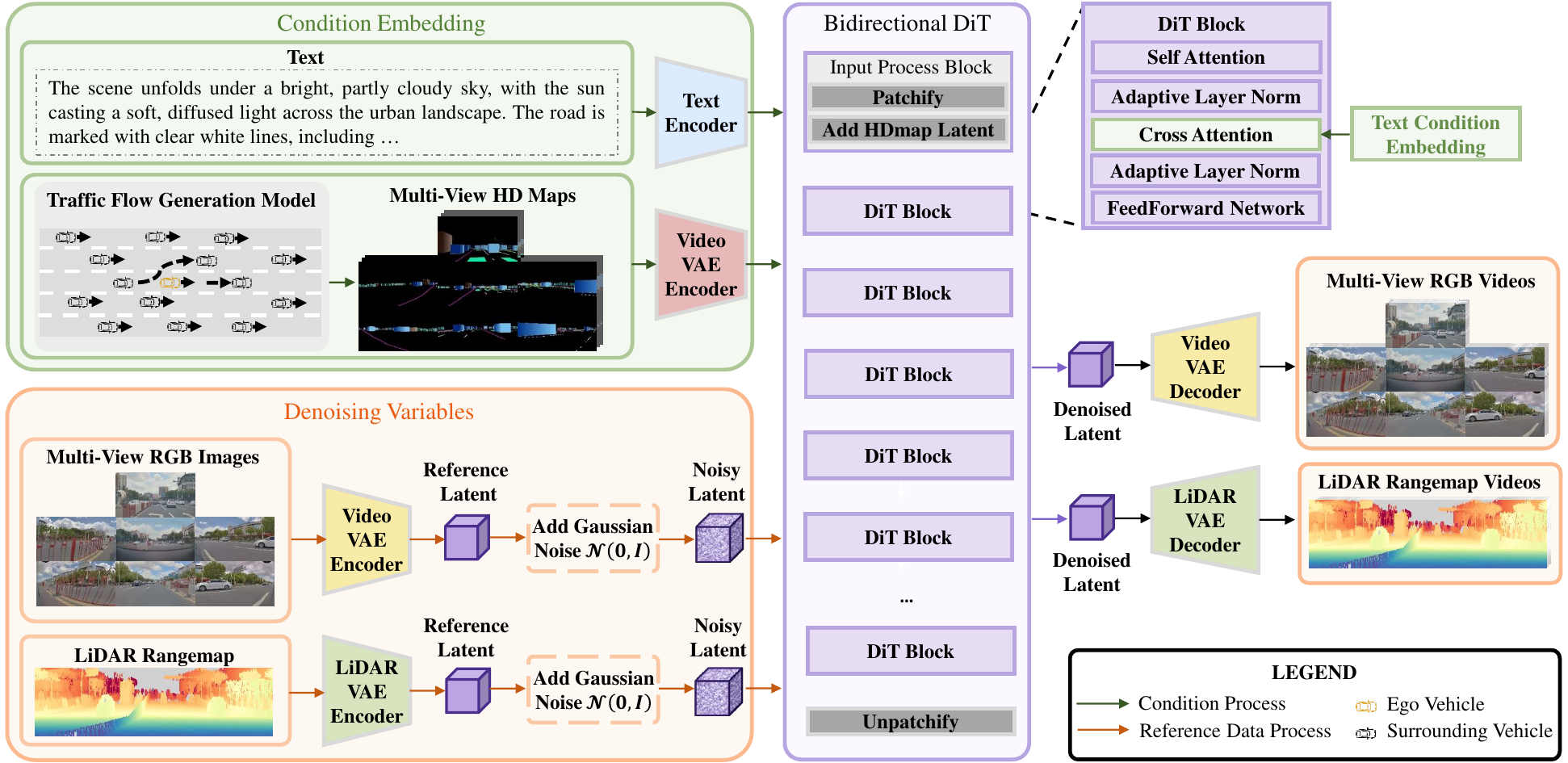}
	\caption{The overview of our DiT-based action-conditioned world model.}
	\label{fig:awm_overview}
    \vspace{-0.3cm}
\end{figure}

\subsubsection{Action-Conditioned World Model}\label{sec:method_wm_arch}
The action-conditioned world model predicts scene transitions after the ego vehicle plus surrounding agents execute specified actions. The model is conditioned on an HD map. To support LiDAR-dependent downstream tasks such as 3DGS scene extrapolation and closed-loop simulation, the model jointly generates RGB videos with LiDAR rangemaps. Concretely, it takes multi-view images, a LiDAR rangemap, and a scene text description as inputs. Multi-view HD-map videos provide the spatial conditions. The model generates future multi-view videos together with LiDAR rangemaps.

Fig.~\ref{fig:awm_overview} shows the architecture of our action-conditioned world model, which is based on the Diffusion Transformer (DiT)~\citep{peebles2023scalable}. Multi-view images, HD maps, LiDAR rangemaps, and text are first encoded into latent representations by their corresponding VAE encoders. Inside the model, HD-map latents are injected into the RGB latents as spatial conditions. RGB latents and LiDAR latents are concatenated along the token dimension. These tokens are then processed by cascaded DiT blocks together with the text latents. Each DiT block applies full self-attention to the RGB and LiDAR latents. This operation encourages multi-view consistency, multimodal alignment, and temporal smoothness. Each block then applies cross-attention between the text latents and the multimodal latents to preserve semantic alignment. Finally, the video VAE decoder and the LiDAR VAE decoder map the latent representations back to pixel space. They produce the generated multi-view videos with LiDAR rangemaps.

\begin{figure}[t]
	\centering
	\includegraphics[width=1\hsize]{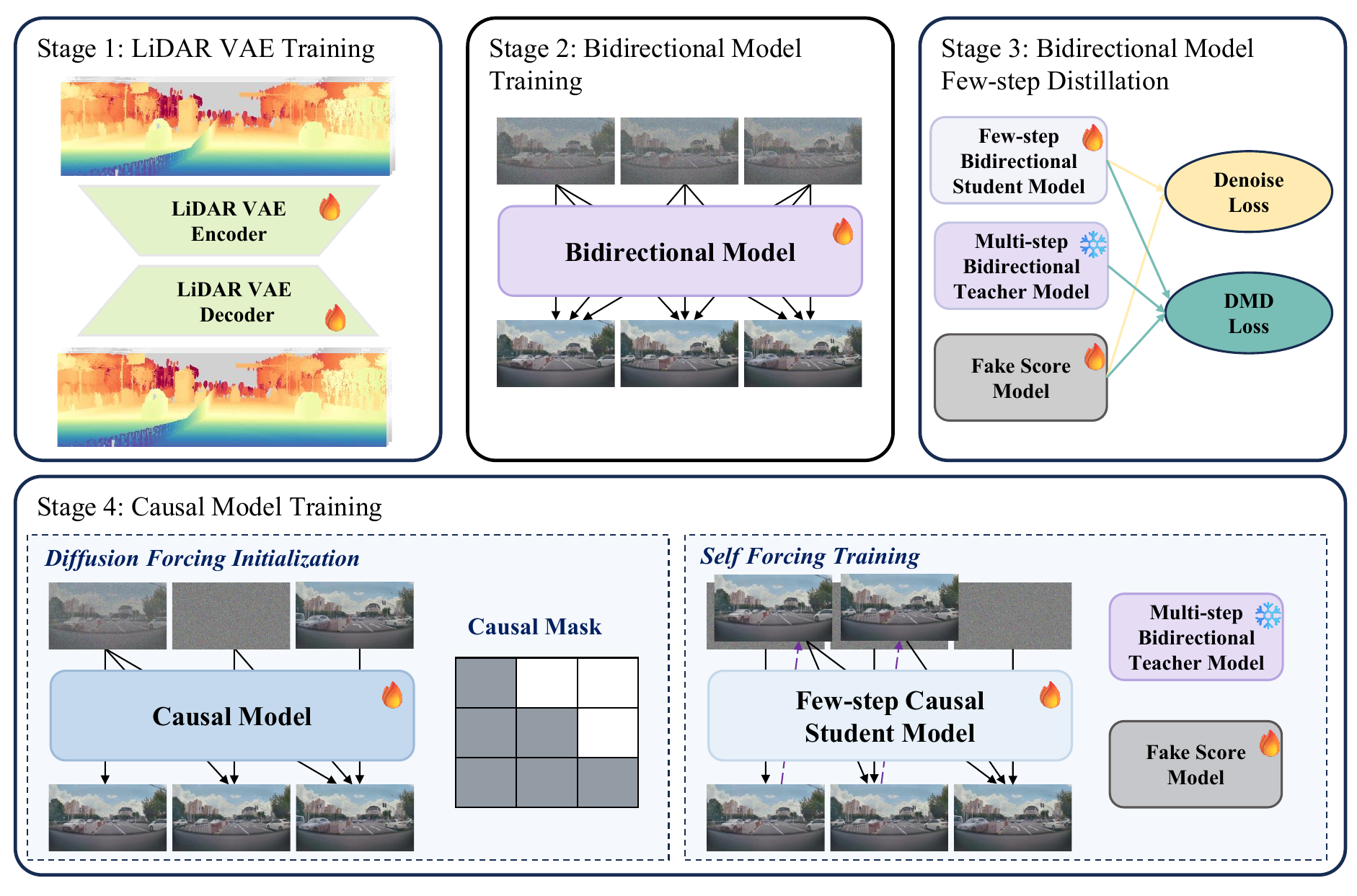}
    \vspace{-0.5cm}
	\caption{Four training stages of the action-conditioned world model.}
	\label{fig:awm_training}
    \vspace{-0.3cm}
\end{figure}

The action-conditioned world model is built upon Cosmos \citep{ali2025world}. To make the model applicable to diverse application scenarios, we develop a four-stage post-training scheme as in Fig. \ref{fig:awm_training}.

\textbf{Stage 1: LiDAR VAE Training}. Although the pretrained video VAE~\citep{wan2025wan} can encode RGB frames together with HD-map frames, it does not model LiDAR rangemaps well. Therefore, we train a separate LiDAR VAE \citep{ren2025cosmos} during the first stage. It uses the same architecture as the video VAE. In detail, we project point clouds from the front LiDAR mounted on our vehicle. The LiDAR has an approximately $120^{\circ}$ field of view. We obtain multiple nine-frame LiDAR-rangemap videos with a resolution of $1280\times128$. Each video sample is resized to $640\times512$ before being passed to the VAE. We repeat the rows by a factor of four, then downsample the columns by a factor of two. The training objective combines the reconstruction loss with the KL-divergence loss.

\textbf{Stage 2: Bidirectional Model Training}. Following \citep{basant2026nvidia}, we train a bidirectional model with strong capability for joint RGB video and LiDAR rangemap generation, which is conditioned on HD maps and text descriptions. For robustness, we use rectified flow ~\citep{liu2023flow} as the training strategy. Let $z_0$ denote the clean latent and $\epsilon \sim \mathcal{N}(0, 1)$ the gaussian noise. The forward diffusion process $f(z_0, t)$ is defined as:
\begin{equation}
	f(z_0, t) = z_t = (1 - t)z_0 + t\epsilon, \quad\quad t \in [0, 1].
\end{equation}
The model is trained to predict the velocity field $v_{\theta}(z_t, t, c)$ with the objective:
\begin{equation}
	\mathcal{L}_{\mathrm{rf}} = \mathbb{E}_{t,\, z_0,\, \epsilon} \left[ \left\| v_\theta(z_t,\, t,\, c) - (\epsilon - z_0 ) \right\|_2^2 \right],
\end{equation}
where $c$ denotes the conditions.

The training resolution for RGB videos is $1280\times720$. The original LiDAR rangemaps have a resolution of $1280\times128$. To align the latent sizes of the two modalities, we resize the rangemaps to $1280\times512$ by repeating rows. We then zero-pad the tail of the LiDAR latents.

\textbf{Stage 3: Bidirectional Model Few-step Distillation}.
Although the bidirectional model from Stage 2 provides strong generation quality, the required multiple denoising steps (usually more than 30 steps) limit its practicality. Therefore, we use DMD2~\citep{yin2024improved} to distill a few-step bidirectional model for efficient data generation.

We treat the model from stage 2 as the teacher model, using its weights to initialize the four-step student model $v_{\phi}$ and fake score model $v_{\mu}$. The student model is trained with the Distribution Matching Distillation (DMD) objective to align the distribution of the few-step student and the multi-step teacher:
\begin{equation}
	\mathcal{L}_{\mathrm{DMD}} = D_{\mathrm{KL}}\left( p_\phi \,\|\, p_{\theta} \right).
	\label{eq:dmd}
\end{equation}
The fake score model is trained with the denoising objective as~\citep{lipman2022flow} to make the fake score model reflect the real-time distribution of the student:
\begin{equation}
	\mathcal{L}_{\mathrm{fake\_score}} = \mathbb{E}_{t,\, \hat{z}_0,\, \epsilon} \left[ \left\| v_\mu(\hat{z}_t,\, t,\, c) - (\epsilon - \hat{z}_0 ) \right\|_2^2 \right],
	\label{eq:fake_score}
\end{equation}
where $\hat{z}_0$ is generated by denoising the student from pure noise, and $\hat{z}_t$ is obtained by applying the forward diffusion process to $\hat{z}_0$. In addition, we update the student model and the fake-score model at a frequency ratio of $1{:}5$ for training stability. 

\textbf{Stage 4: Causal Model Training}. 
The distilled model from Stage 3 accelerates generation while preserving generation quality. However, it does not support real-time autoregressive inference with long-horizon generation. These capabilities are essential for closed-loop simulation. Therefore, we conduct causal model training during the final stage. The training procedure contains two steps.

* \textit{Diffusion Forcing Initialization}. 
We first use diffusion forcing~\citep{chen2024diffusion} to equip the multi-step bidirectional model from Stage 2 with causal inference capability. Unlike bidirectional training in Stage 2, diffusion forcing applies a causal mask. Each token can attend only to tokens that precede it in the temporal sequence. The method also injects different noise levels into different tokens. This design encourages the model to predict the current clean velocity from diverse historical generation states. The training objective is:
\begin{equation}
	\mathcal{L}_{\mathrm{df}} = \mathbb{E}_{t,\, z_0,\, \epsilon} \left[ \left\| v_\theta(z_t^{(i)},\, t,\, c, \tilde{z}^{(<i)}) - (\epsilon - z_0^{(i)} ) \right\|_2^2 \right],
\end{equation}
where $\tilde{z}^{(<i)} = [z_{t_0}^{(0)}, z_{t_1}^{(1)}, \ldots , z_{t_{i - 1}}^{(i - 1)}]$, representing the noisy tokens before the frame $i$.

* \textit{Self Forcing Training}. 
After diffusion forcing training, the model acquires preliminary causal inference capability. It remains a slow multi-step model with limited long-horizon generation quality. Therefore, we further adopt self-forcing~\citep{huang2025self}, which bridges the gap between training with teacher-generated context and inference with the model's own history. In effect, it distills the bidirectional model into a few-step causal model through DMD. We use the bidirectional model from Stage 2 as the teacher model. We initialize the fake-score model with the teacher weights. The student model is initialized from the diffusion-forcing model. The training objectives follow Eq.~\ref{eq:dmd} and Eq.~\ref{eq:fake_score}. The student generates the complete sequence through chunk-by-chunk rollout conditioned on previously generated chunks. For efficiency, we limit the history context to three chunks during training, corresponding to six latent tokens. After training, the model performs autoregressive inference with few denoising steps. It can also generate long sequences suitable for closed-loop simulation.

\subsubsection{3D Gaussian Splatting}\label{sec:method_wm_3dgs}
\textbf{Scene Reconstruction.}
Representing dynamic driving scenes with 3D Gaussian primitives has become a prominent paradigm for autonomous-driving simulation. Recent methods such as Street Gaussians~\citep{yan2024street} and OmniRe~\citep{chen2025omnire} demonstrate high-fidelity reconstruction with novel-view rendering in complex urban environments. Building upon OmniRe, we develop a driving-scene simulation system for closed-loop evaluation. We redesign the road surfaces and dynamic actors to improve geometric consistency and novel-view rendering quality. These changes support reliable closed-loop simulation. Further implementation details are provided in Appendix~\ref{app:3dgs}.

\textbf{Scene Extrapolation.}
Despite strong rendering quality within regions covered by the original observations, reconstruction-based methods remain constrained by a fixed spatial boundary. This limitation becomes particularly problematic when they are deployed as closed-loop world simulators. During a closed-loop rollout, the ego vehicle may deviate from the expert trajectory or enter viewpoints absent from the original log. When the rendering trajectory extends beyond the reconstructed region, the quality of these out-of-boundary (OOB) areas deteriorates rapidly. Typical artifacts include blurred textures, incomplete geometry, and severe floaters.

To address this limitation, we augment the 3D Gaussian reconstruction pipeline with BehaviorFlow and an action-conditioned world model. This combination forms a generative extrapolation framework for future traffic evolution with multimodal sensor observations. BehaviorFlow takes the final observed state of the original log as input. It predicts the subsequent traffic evolution for the ego vehicle and surrounding agents. Conditioned on these extrapolated traffic states and their action controls, the action-conditioned world model synthesizes temporally coherent multi-view camera observations with corresponding LiDAR point clouds. The generated multimodal observations are appended to the original log. They are then used for joint 3D Gaussian reconstruction. This process extends the reconstructed scene beyond the spatial coverage of the original observations. It also enables robust rendering along deviated trajectories with previously unobserved viewpoints.

\subsection{Action Model}\label{sec:method_am}
The action model denotes the policy improved by our loop, mapping observations and navigation intent to future ego‑agent behaviors. We validate it using both a Vision‑Language‑Action (VLA) model and an end‑to‑end planner. For the VLA branch, we employ our previously proposed ChainFlow‑VLA~\citep{wang2026chainflow} (Sec.~\ref{sec:method_am_vla}), a VLA that decomposes planning into autoregressive proposal generation and VLM‑guided residual refinement. For the end-to-end planner, we adopt DiffusionDrive \citep{liao2025diffusiondrive} (Sec.~\ref{sec:method_am_e2e}).

\subsubsection{Vision-Language-Action Model}\label{sec:method_am_vla}
Existing VLA methods for autonomous driving typically follow one of two paradigms: either extending the VLM vocabulary with trajectory tokens and fine-tuning the pretrained model to autoregressively generate driving actions, or using vision-language models (VLMs) as semantic feature extractors whose representations are consumed by downstream action experts.

For example, RecogDrive~\citep{li2025recogdrive} adapts a pretrained VLM to in-domain driving data organized as a sequential cognitive process, thereby endowing the model with both driving scene understanding and trajectory prediction capabilities. However, directly mapping the VLM’s semantic representations to the continuous action space still leads to a modality mismatch: the semantics encoded by the VLM are not fully aligned with the fine-grained dynamics and control requirements of driving. To bridge this gap, RecogDrive introduces a diffusion-based planner that translates the VLM’s latent cognitive representations into smooth and temporally consistent driving trajectories. In contrast, ChainFlow-VLA uses VLM representations to guide trajectory refinement rather than directly generating the final trajectory.

ChainFlow-VLA differs from both paradigms by using VLM representations to guide trajectory refinement rather than directly generate trajectories. Specifically, it decomposes trajectory prediction into two stages: proposal generation and VLM-guided residual refinement. 

In the first stage, a causal autoregressive planner generates a set of kinematically consistent multimodal trajectory proposals. Let $Y_{\mathrm{AR}} = (y_1, \ldots, y_T)$ denote an autoregressively generated trajectory and $\mathcal{O}$ denote the driving observations. The proposal distribution is factorized as:
\begin{equation} 
	p(Y_{\mathrm{AR}} \mid \mathcal{O}) = \prod_{t=1}^{T} p(y_t \mid y_{<t}, \mathcal{O}).
	\label{eq:ar_distribution} 
\end{equation} 

Sampling from this distribution produces $K$ trajectory proposals, denoted by $\{Y_{\mathrm{AR}}^{(k)}\}_{k=1}^{K}$. In the second stage, a residual diffusion model refines each proposal by modeling the corresponding local conditional distribution: 
\begin{equation} p(Y \mid Y_{\mathrm{AR}}^{(k)}, \mathcal{O}),
	\label{eq:conditional_distribution}
\end{equation}

where $Y$ denotes the final trajectory. ChainFlow-VLA extracts a semantic representation $h_{\mathrm{VLM}}$ from the VLM and uses it to guide the proposal-conditioned refinement: 
\begin{equation} p(Y \mid Y_{\mathrm{AR}}^{(k)}, \mathcal{O}) \approx p(Y \mid Y_{\mathrm{AR}}^{(k)}, h_{\mathrm{VLM}}), \label{eq:vlm_conditioning} 
\end{equation} 

where $h_{\mathrm{VLM}}$ encodes the semantic context of the observations and conditions the local refinement distribution associated with each trajectory mode. Combining the multimodal proposal distribution with the proposal-conditioned refinement distributions yields an implicit mixture formulation: 
\begin{equation} p(Y \mid \mathcal{O}) \approx \sum_{k=1}^{K} p(Y \mid Y_{\mathrm{AR}}^{(k)}, h_{\mathrm{VLM}}) p(Y_{\mathrm{AR}}^{(k)} \mid \mathcal{O}), 
	\label{eq:mixture_approximation}
\end{equation}

where each component represents a local distribution centered around one trajectory proposal. In practice, the diffusion model performs refinement in the residual space between each proposal and the ground-truth trajectory.

\subsubsection{End-to-End Planner}\label{sec:method_am_e2e}
DiffusionDrive directly maps raw sensor observations to a distribution over future ego trajectories. Rather than regressing a single deterministic trajectory, it formulates planning as conditional trajectory generation. It employs a truncated diffusion policy to capture the multimodal nature of driving behavior efficiently. Specifically, it constructs a set of prior trajectory anchors by clustering trajectories in the training data. It perturbs these anchors with a truncated noise schedule, which yields an anchored Gaussian distribution. Compared with a conventional diffusion policy initialized from unconstrained Gaussian noise, these anchor-centered initializations provide plausible driving priors while substantially shortening the denoising process.

Given the sampled noisy trajectories, a cascaded diffusion decoder progressively refines them under the current scene representation. The decoder interacts with BEV or perspective-view features through deformable attention. It further incorporates structured agent and map queries produced by the perception module. At each denoising step, it predicts refined trajectory coordinates and their confidence scores. The trajectory with the highest confidence is selected as the final ego plan. This design preserves the diversity of diffusion-based trajectory generation while requiring only a few denoising steps, which makes DiffusionDrive suitable for real-time end-to-end planning.

\subsection{BehaviorFlow}\label{sec:behaviorflow}
BehaviorFlow is the structured traffic-generation module that connects the action model to the world simulator within BehaviorWorldGen. Inspired by \citep{peng2025infgen} and our prior work \citep{zhao2025autoregressive}, it models the evolving traffic scene as an autoregressive sequence over agent lifecycles, frame-level meta-actions, and agent states. BehaviorFlow supports explicit control over scene composition and agent behavior. All unconstrained variables are generated jointly under these interventions, producing interaction-consistent multi-agent rollouts for world simulation.

\paragraph{Behavior flow generation.}
Let $\mathcal{S}$ denote the static scene context, including the HD map, traffic lights, traffic signs, and other road infrastructure. These scene elements are provided as conditions for traffic generation.

For each agent $i$, let $\ell_{i,t}$ denote its lifecycle variable at timestep $t$. We use $\mathbf{L}_t=\{\ell_{i,t}\}_{i\in\mathcal{I}_t}$ to denote the lifecycle variables of all candidate agents, where $\mathcal{I}_t$ contains the agents inherited from the preceding timestep together with candidate agents that may enter the scene. For every active agent $i\in\mathcal{I}_t$, let $\alpha_{i,t}$ denote its frame-level meta-action and $\mathbf{x}_{i,t}$ its physical state. Their corresponding collections are $\mathbf{A}_t=\{\alpha_{i,t}\}_{i\in\mathcal{I}_t}$ and $\mathbf{X}_t=\{\mathbf{x}_{i,t}\}_{i\in\mathcal{I}_t}$, respectively.
Here, $\mathbf{x}_{i,t}$ describes the agent category, shape, pose, heading, and velocity.

Let $\mathcal{H}_{<t}=\left(\mathbf{L}_{<t},\mathbf{A}_{<t},\mathbf{X}_{<t}\right)$ denote the traffic history available before timestep $t$. A generated rollout is represented as
\begin{equation}
	\mathcal{R}_{1:T}
	=
	\left\{
	\left(
	\mathbf{L}_t,
	\mathbf{A}_t,
	\mathbf{X}_t
	\right)
	\right\}_{t=1}^{T}.
\end{equation}
BehaviorFlow can be formulated as
\begin{equation}
	\begin{aligned}
		p\!\left(
		\mathcal{R}_{1:T}
		\mid
		\mathcal{S},\mathcal{H}_{<1}
		\right)
		=
		\prod_{t=1}^{T}
		&p\!\left(
		\mathbf{L}_t
		\mid
		\mathcal{S},\mathcal{H}_{<t}
		\right)
		\\
		&\cdot p\!\left(
		\mathbf{A}_t
		\mid
		\mathbf{L}_t,
		\mathcal{S},\mathcal{H}_{<t}
		\right)
		\\
		&\cdot p\!\left(
		\mathbf{X}_t
		\mid
		\mathbf{A}_t,\mathbf{L}_t,
		\mathcal{S},\mathcal{H}_{<t}
		\right).
		\label{eq:behaviorflow_factorization}
	\end{aligned}
\end{equation}

\paragraph{Identity-aware lifecycle generation.}
The lifecycle variable of each identity is defined as $\ell_{i,t}\in\{\mathtt{ENTER},\mathtt{ALIVE},\mathtt{EXIT}\}$. Here, $\mathtt{ENTER}$ introduces a new agent into the generated scene. The token $\mathtt{ALIVE}$ preserves an existing agent, whereas $\mathtt{EXIT}$ removes an agent from the subsequent rollout. This construction preserves temporal identity correspondence while allowing the number and composition of traffic participants to evolve over time.

The lifecycle variables are decoded before any behavioral or physical-state attributes:
\begin{equation}
	\mathbf{L}_t
	\sim
	p\!\left(
	\mathbf{L}_t
	\mid
	\mathcal{S},\mathcal{H}_{<t}
	\right).
	\label{eq:lifecycle_generation}
\end{equation}
The lifecycle factor first determines which agents participate in the current interaction. The behavior and state factors then operate only on the resulting active agents.

\paragraph{Meta-action-conditioned agent generation.}
For each active agent, $\alpha_{i,t}$ denotes the high-level behavior governing its transition into state $\mathbf{x}_{i,t}$. We use the frame-level meta-action vocabulary
\begin{equation}
	\begin{split}
		\mathcal{A}_{\mathrm{meta}}=\{&
		\mathtt{Stationary},
		\mathtt{LaneChangeLeft},
		\mathtt{LaneChangeRight},
		\mathtt{TurnLeft},
		\mathtt{TurnRight},\\
		&
		\mathtt{UTurnLeft},
		\mathtt{UTurnRight},
		\mathtt{BypassLeft},
		\mathtt{BypassRight},
		\mathtt{KeepStraight}
		\}.
	\end{split}
	\label{eq:meta_action_vocabulary}
\end{equation}
Unlike a single behavior label assigned to an entire prediction horizon, frame-level meta-actions remain temporally aligned with the corresponding state transitions. The meta-actions of all active agents are generated jointly as
\begin{equation}
	\mathbf{A}_t
	\sim
	p\!\left(
	\mathbf{A}_t
	\mid
	\mathbf{L}_t,\mathcal{S},\mathcal{H}_{<t}
	\right).
	\label{eq:meta_action_generation}
\end{equation}

Conditioned on the lifecycle and meta-action variables, BehaviorFlow directly generates the physical states of all active agents:
\begin{equation}
	\mathbf{X}_t
	\sim
	p\!\left(
	\mathbf{X}_t
	\mid
	\mathbf{A}_t,\mathbf{L}_t,
	\mathcal{S},\mathcal{H}_{<t}
	\right).
	\label{eq:agent_state_generation}
\end{equation}
For agent $i$, we parameterize its state as
\begin{equation}
	\mathbf{x}_{i,t}
	=
	\left(
	\tau_{i,t},
	\mathbf{d}_{i,t},
	m_{i,t},
	\mathbf{r}_{i,t}
	\right),
\end{equation}
where $\tau_{i,t}$ denotes the agent category, $\mathbf{d}_{i,t}$ describes its physical dimensions, $m_{i,t}$ is the associated map-segment index, and $\mathbf{r}_{i,t}$ represents its position, heading, and velocity relative to that segment.

Let $\mathbf{c}_{i,t}$ collect the scene context, traffic history, current lifecycle variables, and current meta-actions. It also includes the state tokens already decoded at timestep $t$. The state distribution for an active agent can be written as
\begin{equation}
	\begin{aligned}
		&p\!\left(
		\mathbf{x}_{i,t}
		\mid
		\mathbf{c}_{i,t}
		\right)
		\\
		&=
		\begin{cases}
			p\!\left(
			\tau_{i,t},\mathbf{d}_{i,t}
			\mid
			\mathbf{c}_{i,t}
			\right)
			p\!\left(
			m_{i,t}
			\mid
			\tau_{i,t},\mathbf{d}_{i,t},\mathbf{c}_{i,t}
			\right)
			\\ \qquad\cdot
			p\!\left(
			\mathbf{r}_{i,t}
			\mid
			m_{i,t},\tau_{i,t},\mathbf{d}_{i,t},\mathbf{c}_{i,t}
			\right),
			&
			\ell_{i,t}=\mathtt{ENTER},
			\\[6pt]
			\mathds{1}\!\left[
			(\tau_{i,t},\mathbf{d}_{i,t})
			=
			(\tau_{i,t-1},\mathbf{d}_{i,t-1})
			\right]
			p\!\left(
			m_{i,t},\mathbf{r}_{i,t}
			\mid
			\mathbf{c}_{i,t}
			\right),
			&
			\ell_{i,t}=\mathtt{ALIVE}.
		\end{cases}
		\label{eq:conditional_agent_state}
	\end{aligned}
\end{equation}
For a newly entering agent, the model generates its time-invariant attributes together with its initial dynamic state. For an existing agent, its category and shape remain unchanged. Its map association, pose, heading, and velocity are updated autoregressively. Since the state factor is jointly conditioned on the lifecycle and meta-actions of the active population, an injected behavior can alter the controlled agent as well as the reactions of surrounding agents.

\paragraph{Training objective.}
Lifecycle labels are obtained from identity validity in the driving logs. The first valid frame of an identity is labeled $\mathtt{ENTER}$, subsequent valid frames are labeled $\mathtt{ALIVE}$, and its transition out of the scene is labeled $\mathtt{EXIT}$. Frame-level meta-actions are inferred from consecutive agent states under the map topology. Agent attributes and dynamic states are discretized into categorical tokens following the map-relative representation described above.

The training objective follows the three factors in Eq.~\eqref{eq:behaviorflow_factorization}:
\begin{equation}
	\mathcal{L}_{\mathrm{flow}}
	=
	\lambda_{\mathrm{life}}
	\mathcal{L}_{\mathrm{life}}
	+
	\lambda_{\mathrm{meta}}
	\mathcal{L}_{\mathrm{meta}}
	+
	\lambda_{\mathrm{state}}
	\mathcal{L}_{\mathrm{state}},
	\label{eq:behaviorflow_loss}
\end{equation}
where $\mathcal{L}_{\mathrm{life}}$, $\mathcal{L}_{\mathrm{meta}}$, and $\mathcal{L}_{\mathrm{state}}$ are the supervised losses for lifecycle variables, frame-level meta-actions, and agent states, respectively. The coefficients $\lambda_{\mathrm{life}}$, $\lambda_{\mathrm{meta}}$, and $\lambda_{\mathrm{state}}$ balance their contributions. During training, ground-truth lifecycle variables are teacher-forced when predicting meta-actions. Ground-truth lifecycle variables and meta-action variables are then teacher-forced when predicting agent states.

\paragraph{Controllable traffic-flow editing.}
At inference time, lifecycle and meta-action variables can be specified for arbitrary agent--timestep pairs. Let $\Omega_{\mathrm{life}}$ denote the lifecycle constraint set. Let $\Omega_{\mathrm{meta}}$ denote the meta-action constraint set. The controllably decoded lifecycle variables are
\begin{equation}
	\widetilde{\ell}_{i,t}
	=
	\begin{cases}
		\ell^{\star}_{i,t},
		&
		(i,t)\in\Omega_{\mathrm{life}},
		\\[2pt]
		\ell_{i,t}
		\sim
		p\!\left(
		\ell_{i,t}
		\mid
		\mathcal{S},\mathcal{H}_{<t}
		\right),
		&
		\text{otherwise},
	\end{cases}
	\label{eq:lifecycle_editing}
\end{equation}
and the controllably decoded meta-actions are
\begin{equation}
	\widetilde{\alpha}_{i,t}
	=
	\begin{cases}
		\alpha^{\star}_{i,t},
		&
		(i,t)\in\Omega_{\mathrm{meta}},
		\\[2pt]
		\alpha_{i,t}
		\sim
		p\!\left(
		\alpha_{i,t}
		\mid
		\widetilde{\mathbf{L}}_t,
		\mathcal{S},\mathcal{H}_{<t}
		\right),
		&
		\text{otherwise}.
	\end{cases}
	\label{eq:meta_action_editing}
\end{equation}
After these interventions, the complete agent state is regenerated according to
\begin{equation}
	\widetilde{\mathbf{X}}_t
	\sim
	p\!\left(
	\mathbf{X}_t
	\mid
	\widetilde{\mathbf{A}}_t,
	\widetilde{\mathbf{L}}_t,
	\mathcal{S},\mathcal{H}_{<t}
	\right).
	\label{eq:controlled_state_generation}
\end{equation}
A lifecycle intervention changes the traffic composition, whereas a meta-action intervention changes the intended behavior. In both cases, the state factor regenerates all active agents jointly. This joint generation allows unconstrained agents to respond to the modified scene.

BehaviorFlow is applied to scenarios in which the current action model performs poorly, such as collisions, unsafe gap selection, failed yielding, or insufficient route progress. Based on the diagnosed failure modes, lifecycle constraints modify the traffic participants. Meta-action constraints specify their intended behaviors. These controllable edits produce diverse interaction-consistent traffic scenes, including alternative agent compositions and behavioral interactions that are difficult to obtain from passive driving logs.

Each generated rollout is combined with the scene context $\mathcal{S}$ to construct multi-view layouts for the world simulator. The world simulator renders these layouts into realistic sensor observations. These observations are paired with the corresponding interaction-aware trajectories before being added to the training data of the action model. In this way, BehaviorFlow transforms identified action-model failures into diverse, controllable training examples for subsequent policy refinement.

\section{Experiments}
\subsection{World Simulator Results}

\subsubsection{Action-Conditioned World Model}\label{sec:exp_awm}
We evaluate the action-conditioned world model on our private dataset in terms of short-horizon video generation, long-horizon video generation, and LiDAR generation. These experiments assess its capability of generating realistic videos and LiDAR point clouds under structured HD-map conditions.

\begin{figure}[p]
	\centering
    \vspace{-0.3cm}
	\includegraphics[width=\hsize]{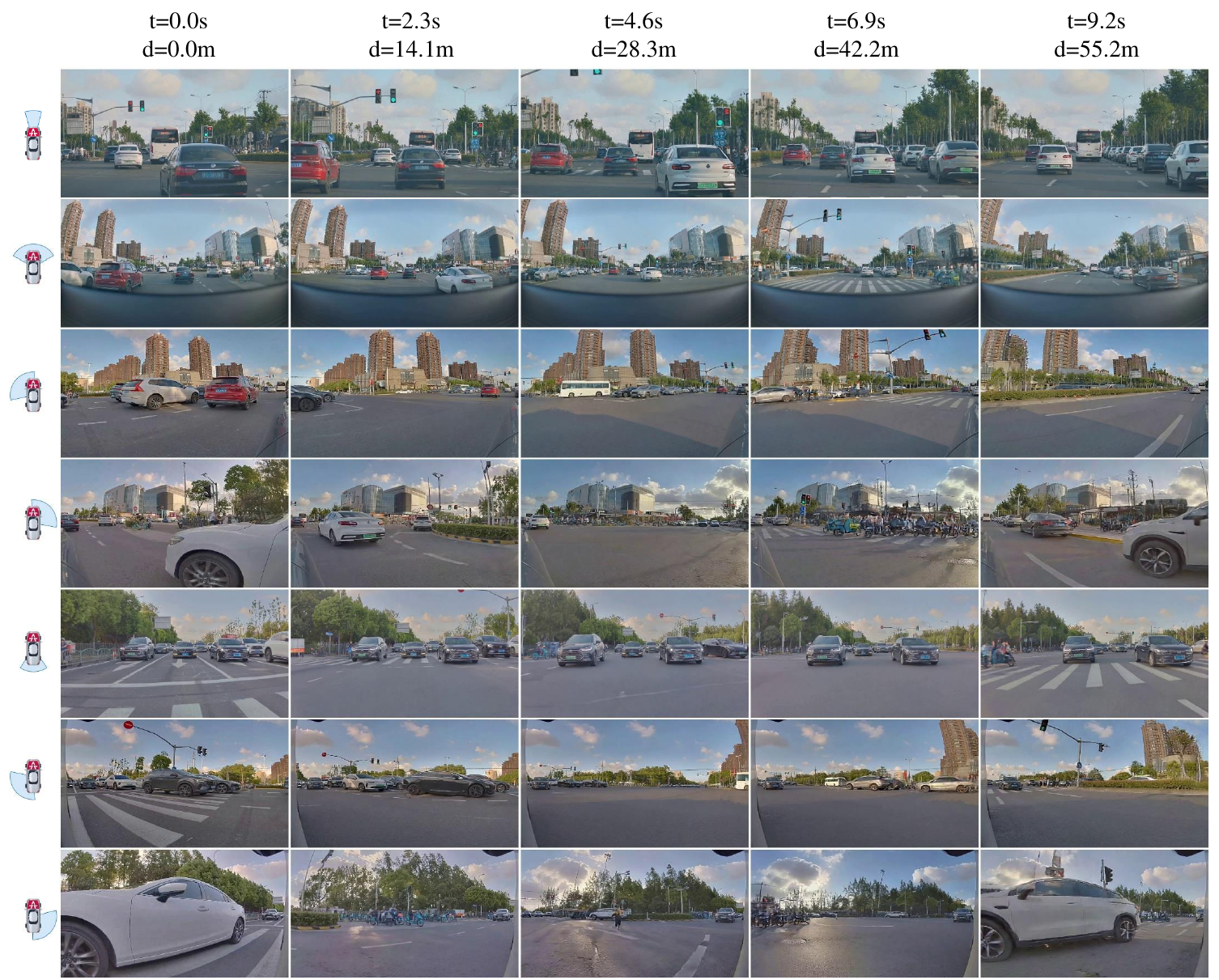}
	\includegraphics[width=\hsize]{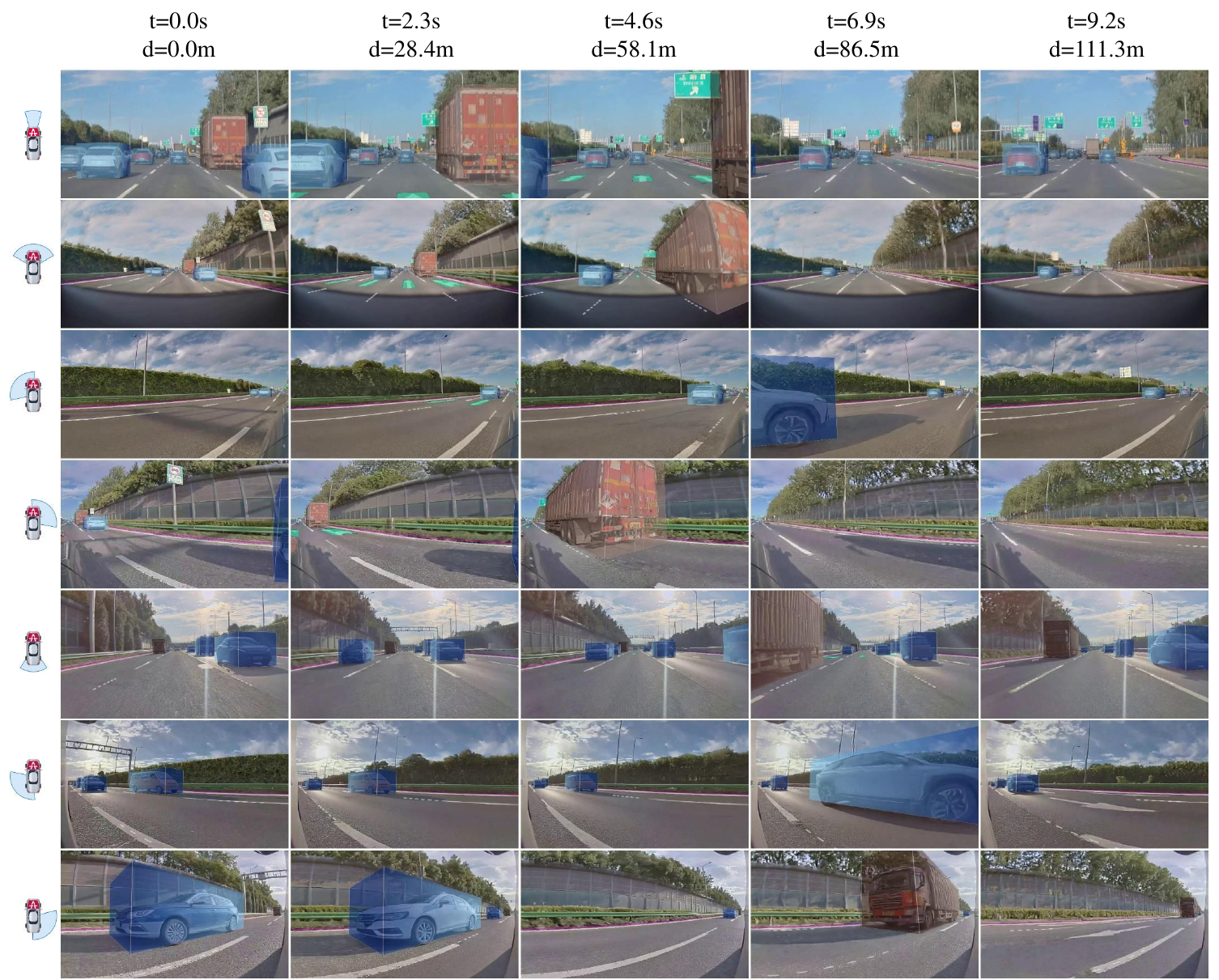}
	\caption{The 93-frame generation results of our action-conditioned world model. Top: RGB. Bottom: RGB with HD-map overlaid.}
	\label{fig:awm-93frame}
\end{figure}

\textbf{Short-time Video Generation.} First, we evaluate the generation quality of our bidirectional model (stage 2) and causal model (stage 4) on the task of short-time video generation (93 Frames, $10$ FPS). As shown in Tab. \ref{tab:awm1}, the causal model achieves better FID/FVD and nearly 70$\times$ acceleration compared with the bidirectional model. These results show that the causal model improves generation quality while substantially increasing inference efficiency. Fig.~\ref{fig:awm-93frame} further shows that its generated videos remain temporally coherent across views. The generated scene content also follows the HD-map conditions closely. 

\begin{figure}[p]
	\centering
    \vspace{-0.3cm}
	\includegraphics[width=\hsize]{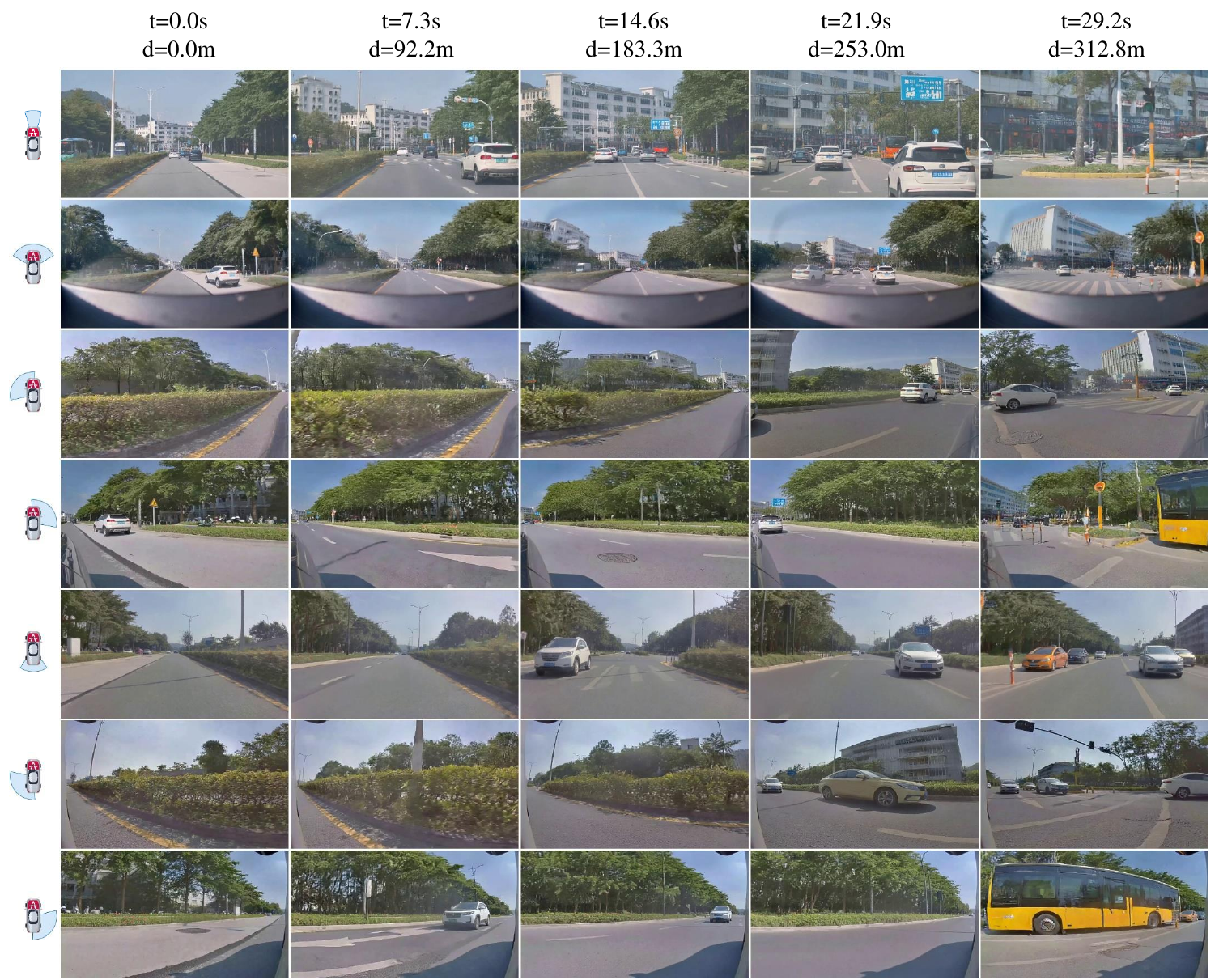}
	\includegraphics[width=\hsize]{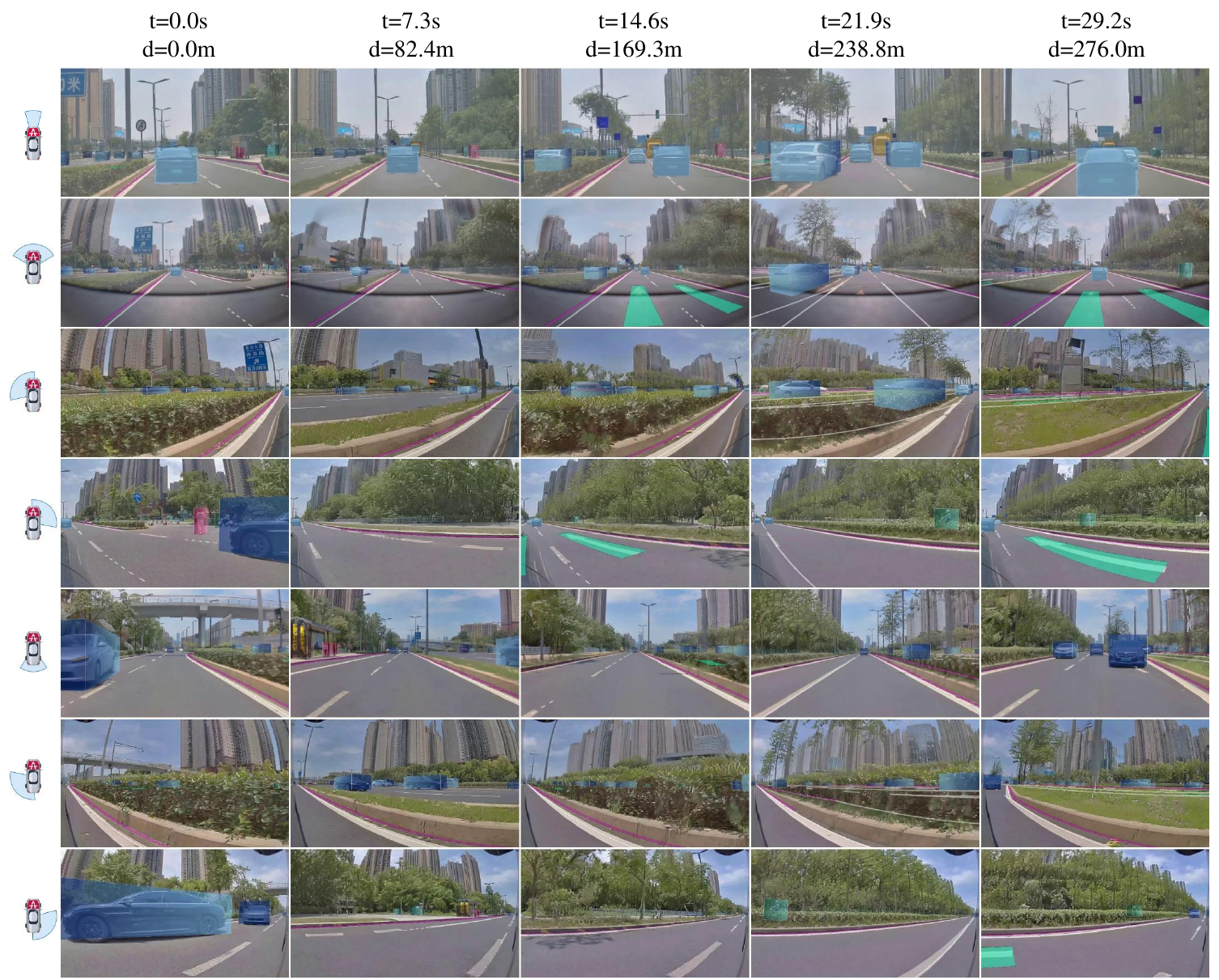}
	\caption{The 30-second generation results of our action-conditioned world model. Top: RGB. Bottom: RGB with HD-map overlaid.}
	\label{fig:awm-300frame}
\end{figure}

\textbf{Long-time Video Generation.} We further evaluate the causal model through autoregressive generation over 30 seconds with a rolling KV cache. As shown in Fig.~\ref{fig:awm-300frame}, the generated videos maintain stable visual quality throughout the rollout. Their scene content also evolves plausibly under the given HD-map conditions. This long-horizon stability indicates that the model can provide reliable visual observations for closed-loop simulation.

\begin{figure}[p]
	\centering
	\includegraphics[width=\hsize]{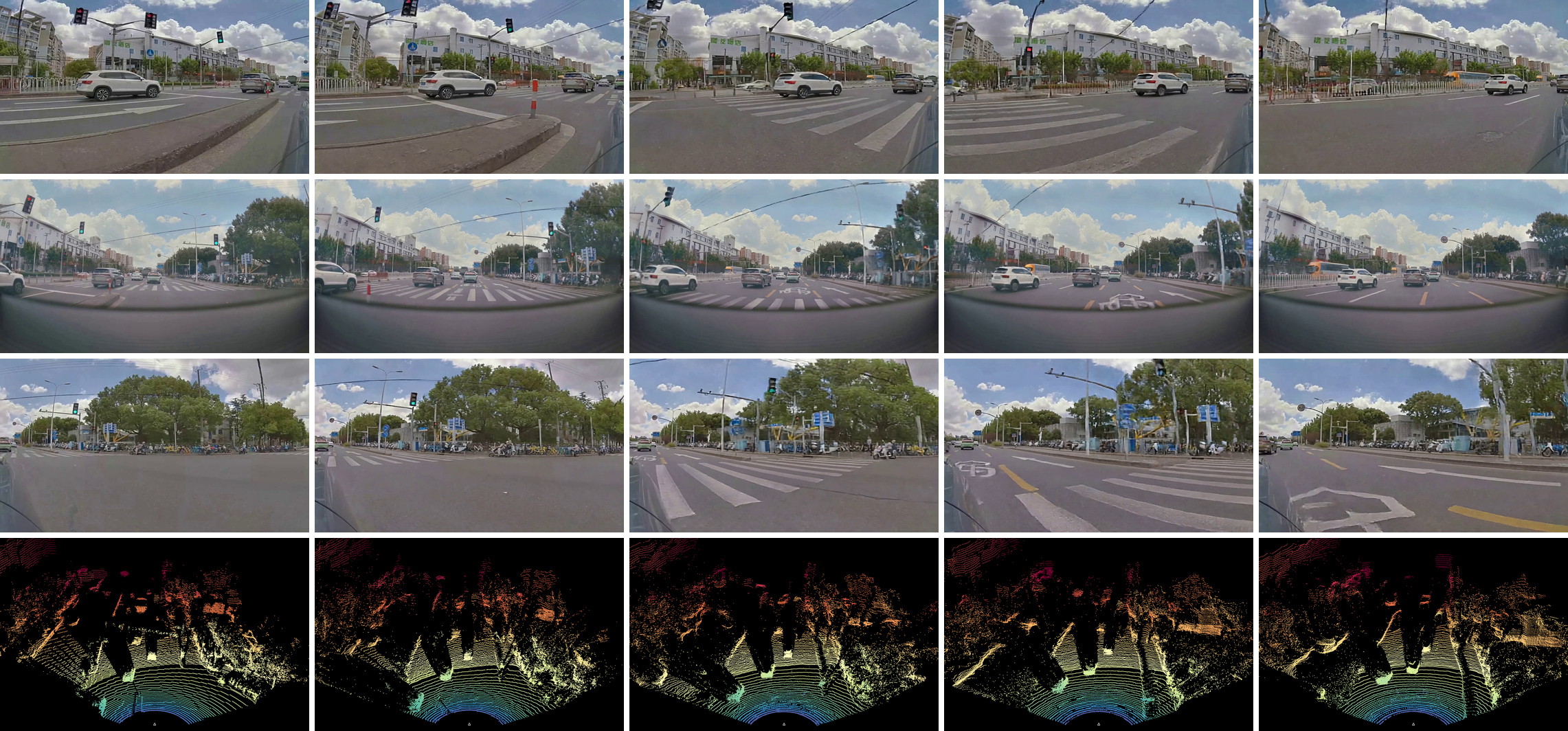}
	\includegraphics[width=\hsize]{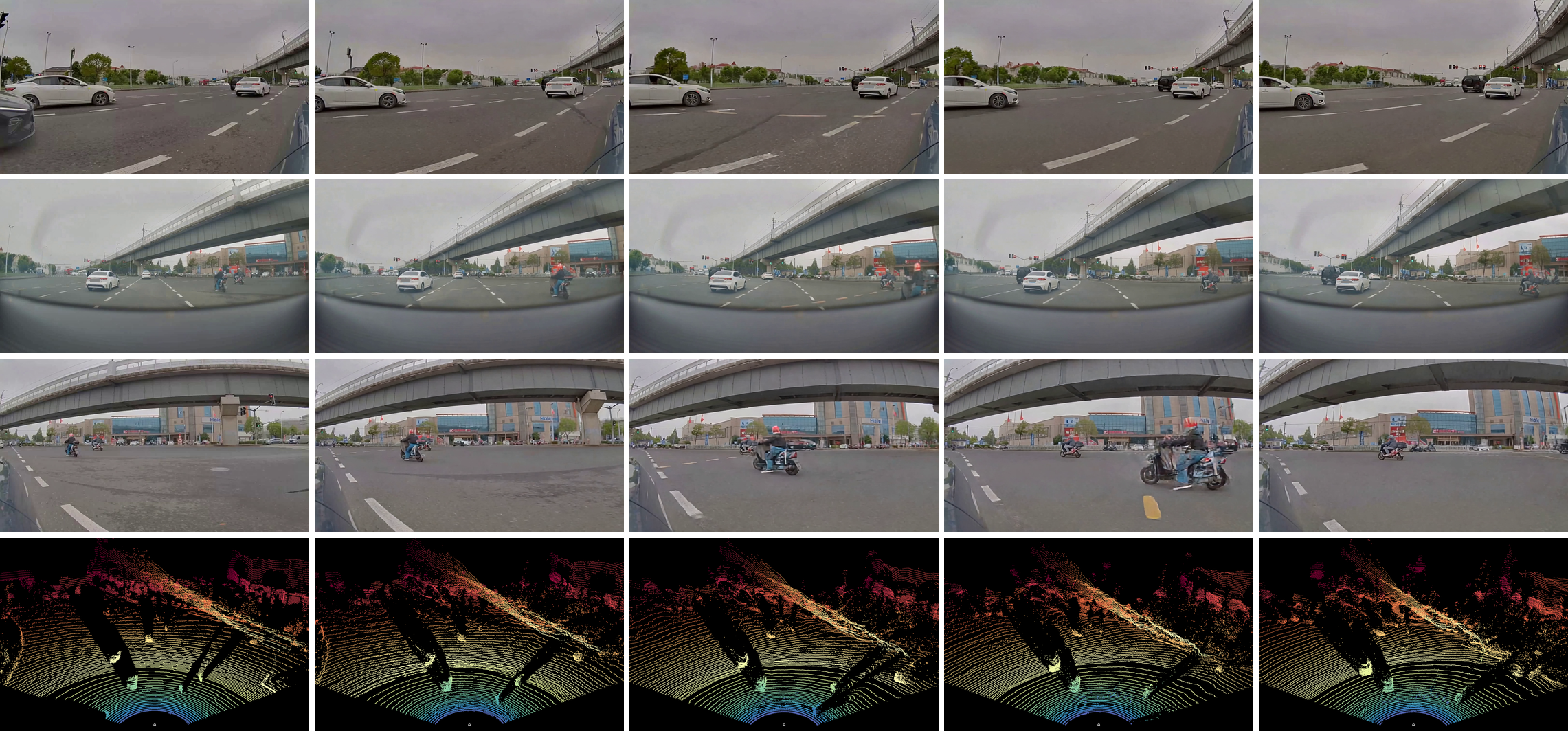}
	\caption{The LiDAR generation results of our action-conditioned world model.}
	\label{fig:awm-lidar}
\end{figure}

\textbf{LiDAR Generation.} We also evaluate the LiDAR generation capability of the model. As shown in Fig.~\ref{fig:awm-lidar}, the generated LiDAR point clouds remain temporally consistent across frames. They also preserve clear structures for the static background and moving vehicles. These results demonstrate that the model can generate geometrically meaningful observations for LiDAR-dependent applications such as 3DGS scene extrapolation.

Beyond standard video and LiDAR generation tasks, we further evaluate the model's capability for controllable scene-level editing through rare-object insertion and environmental appearance control.

\begin{figure}[p]
	\centering
	\includegraphics[width=\hsize]{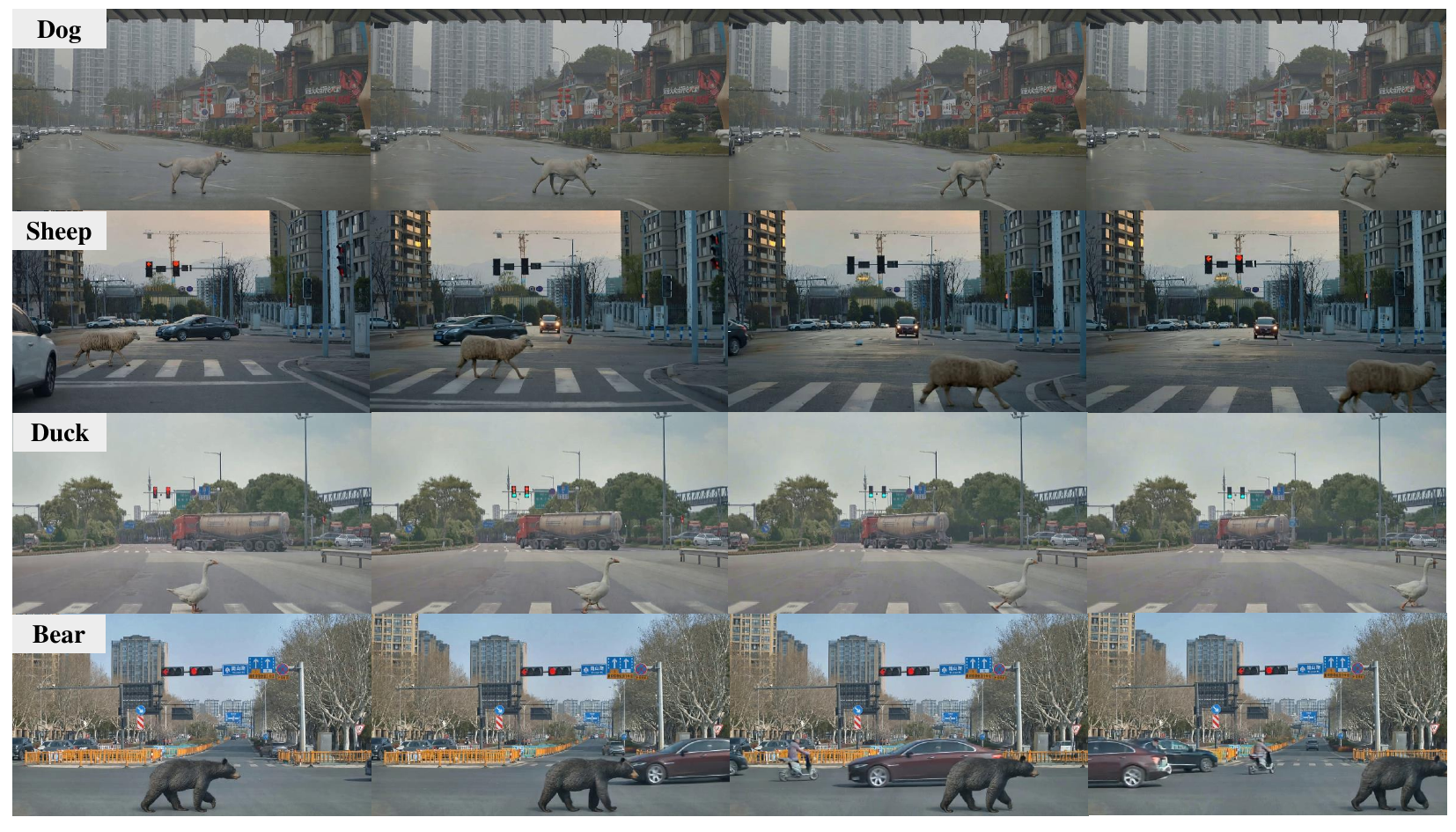}
	\caption{Animal scene generation results of our action-conditioned world model.}
	\label{fig:awm-animal}
\end{figure}

\textbf{Animal Scene Generation.} We first evaluate the model's ability to synthesize rare and safety-critical scene content by inserting animals into driving scenes. As shown in Fig.~\ref{fig:awm-animal}, the inserted animals maintain consistent appearances across frames and exhibit temporally coherent, physically plausible motion. Meanwhile, the surrounding traffic, road layout, and other scene elements remain stable throughout the generation. These results demonstrate that the model can introduce uncommon objects into driving environments without disrupting the overall scene coherence.

\begin{table}[t]
  \caption{Generation speed and quality of the bidirectional model and the causal model.}
  \vspace{-0.3cm}
  \label{tab:awm1}
  \centering
  \begin{tabular}{cccccc}
  \toprule
                      \textbf{Model} & \textbf{Frames} & \textbf{FPS} & \textbf{Speed (on 8$\times$H100 GPUs)} & \textbf{FID}  & \textbf{FVD}   \\
  \midrule
  Bidirectional Model & 93 & 10    & $\sim$20s / frame            & 6.72 & 61.29 \\
  Causal Model        & 93  & 10   & $\sim$0.3s / frame           & 4.36 & 43.27 \\
  \bottomrule
  \end{tabular}
\end{table}

\begin{figure}[p]
	\centering
	\includegraphics[width=\hsize]{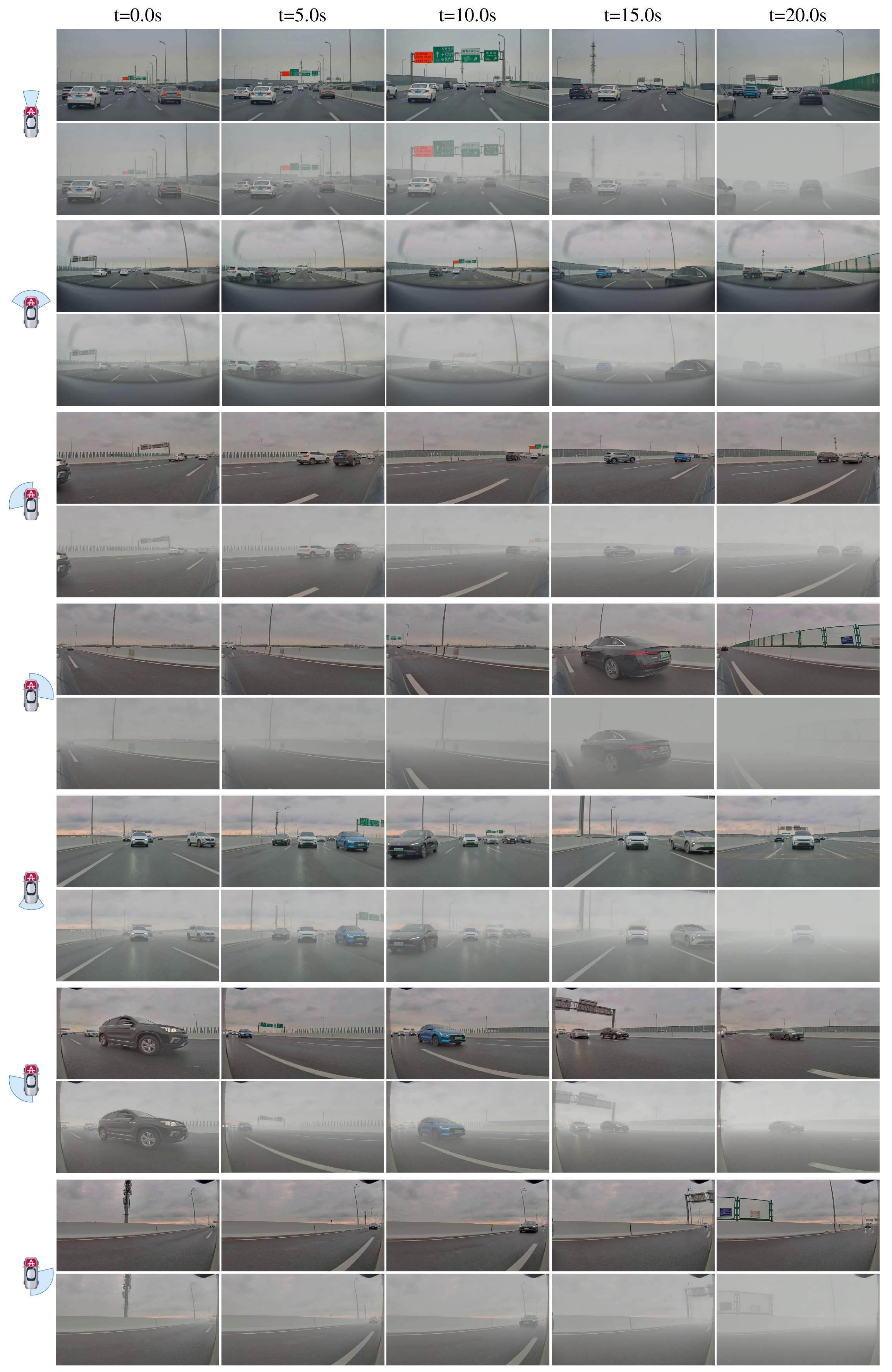}
	\caption{Fog density control results of our action-conditioned world model.}
	\label{fig:awm-fog}
\end{figure}

\textbf{Fog Density Control.} We further assess fine-grained environmental control by progressively increasing the fog density within the same driving scene. As shown in Fig.~\ref{fig:awm-fog}, visibility and contrast decrease smoothly as the fog becomes denser, while the road layout and scene content remain consistent. These results demonstrate that the model supports continuous control over atmospheric conditions beyond discrete weather-style transfer.

\begin{figure}[!t]
	\centering
	\includegraphics[width=\hsize]{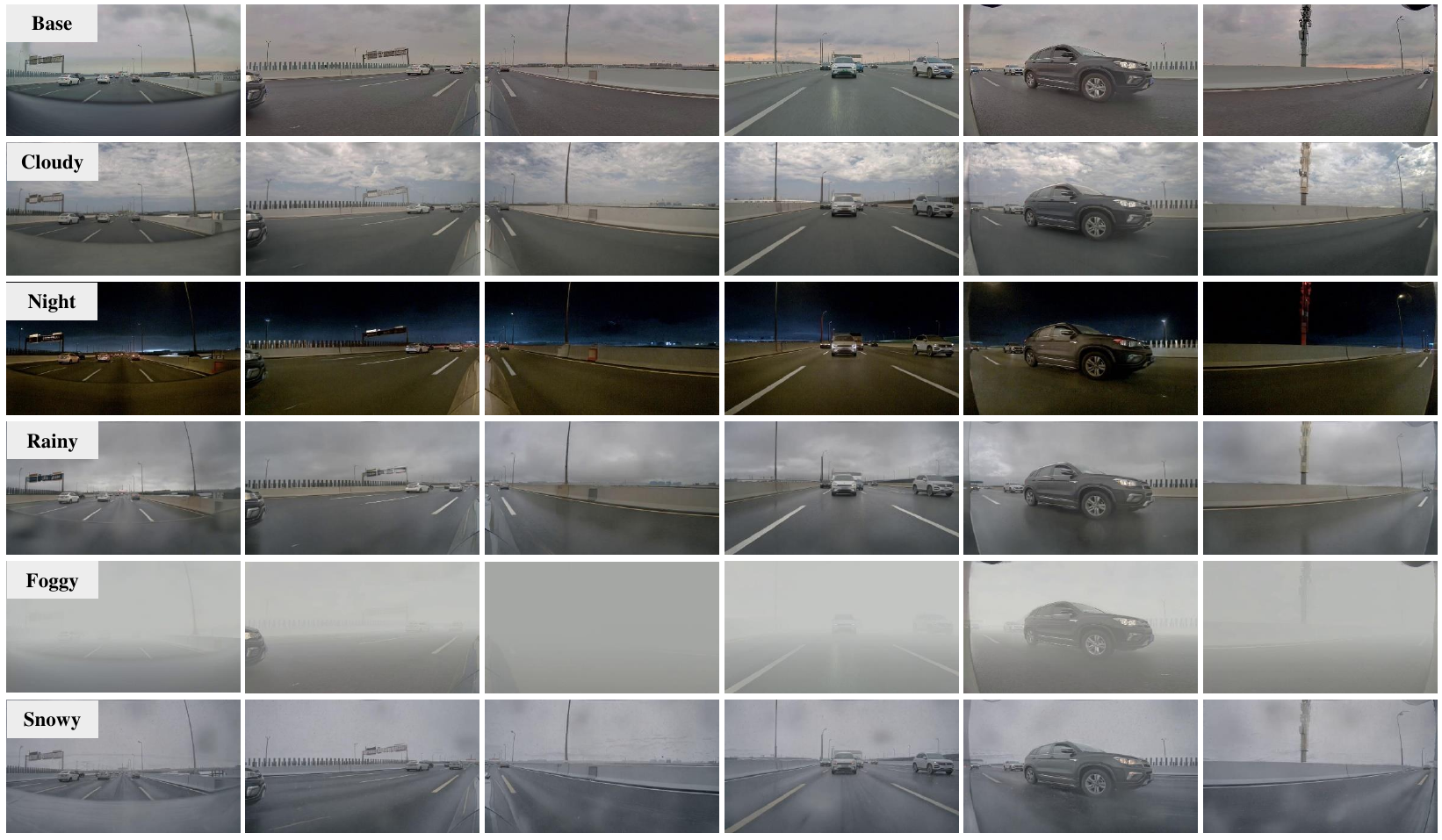}
	\caption{Weather style transfer results of our action-conditioned world model.}
	\label{fig:awm-weather}
\end{figure}

\textbf{Weather Style Transfer.} We next evaluate the model's ability to transfer the same driving scene to diverse environmental conditions. As shown in Fig.~\ref{fig:awm-weather}, the model generates cloudy, nighttime, rainy, foggy, and snowy variants. Across these variants, the underlying road geometry, scene layout, and object identities remain consistent. These results demonstrate that the model can diversify environmental conditions while preserving the semantic and structural content of the original scene.

\clearpage

\clearpage

\begin{figure}[!t]
	\centering
	\includegraphics[width=\hsize]{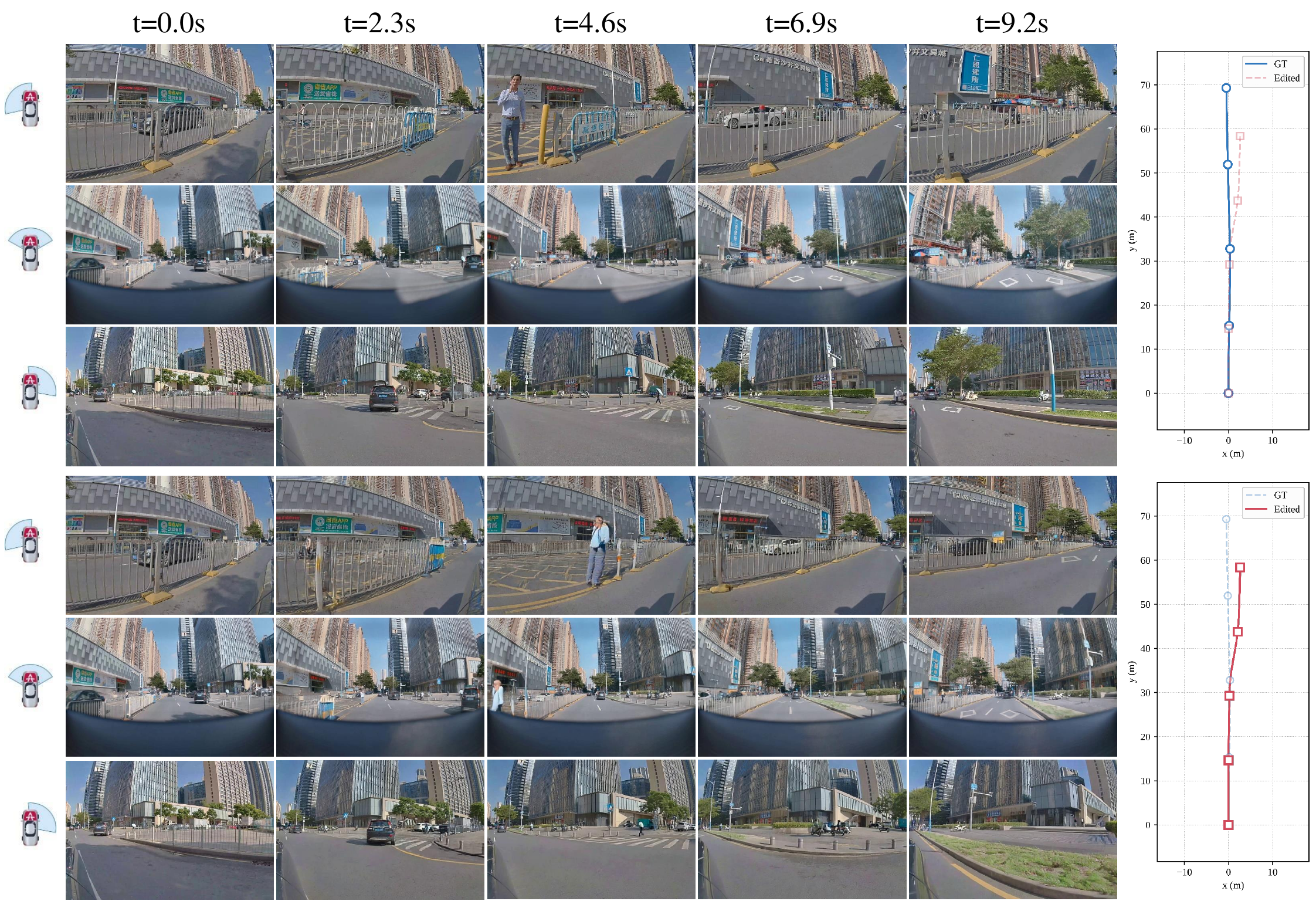}
	\caption{Lane-change trajectory editing results.}
	\label{fig:awm-lanechange}
\end{figure}
	
\begin{figure}[!t]
	\centering
	\includegraphics[width=\hsize]{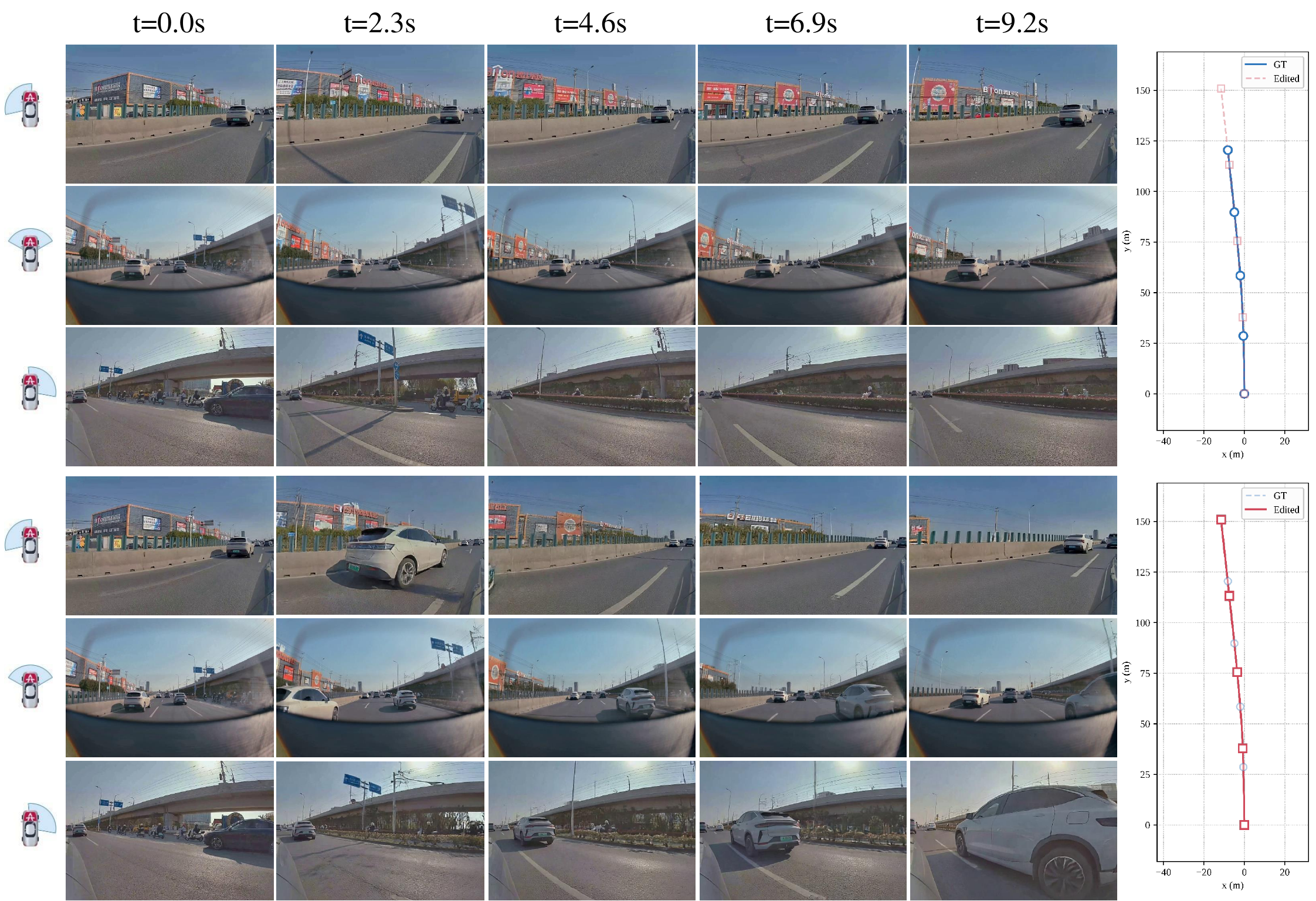}
	\caption{Overtaking trajectory editing results.}
	\label{fig:awm-overtake}
\end{figure}
	
\begin{figure}[!t]
	\centering
	\includegraphics[width=\hsize]{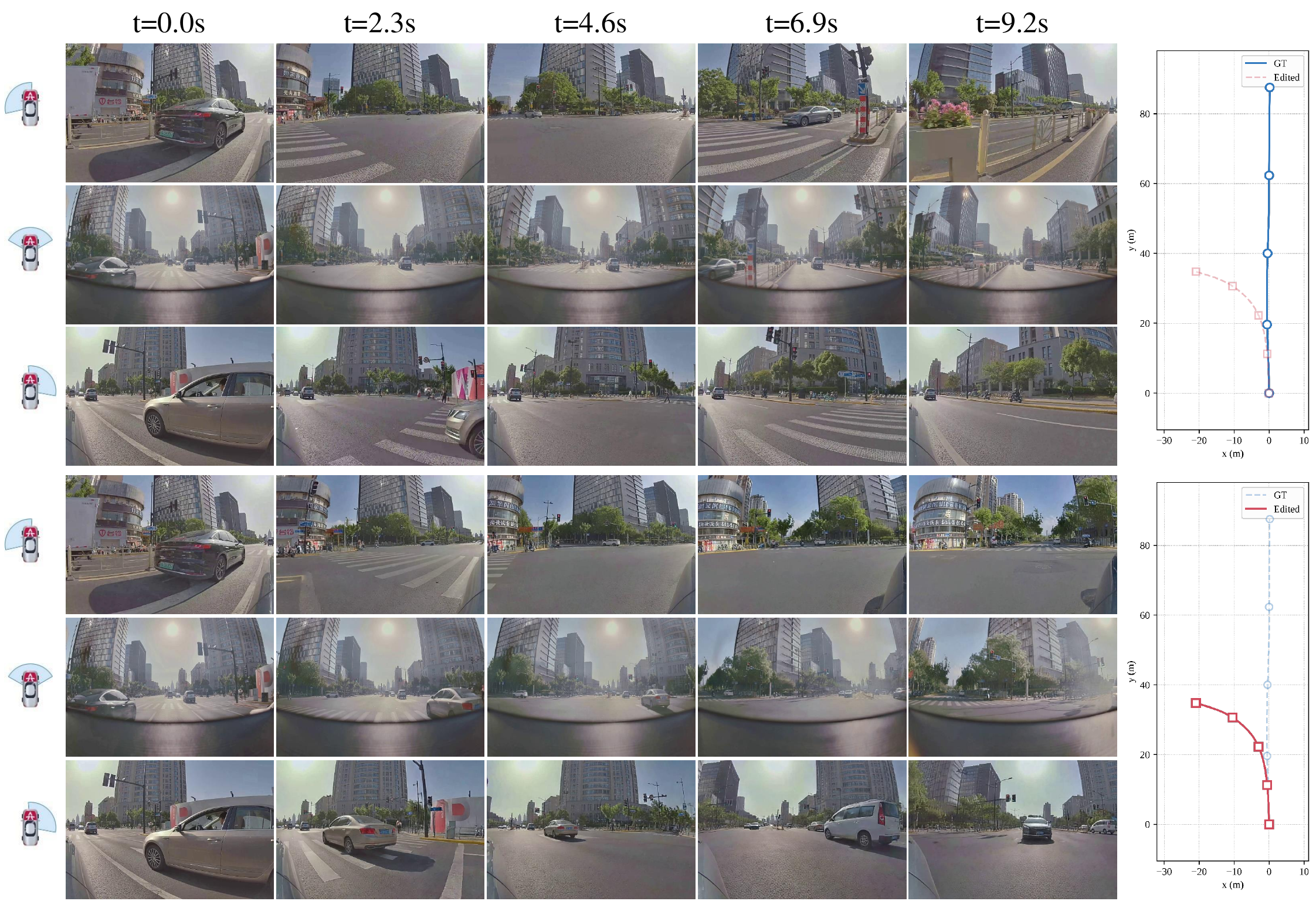}
	\caption{Straight-to-turn trajectory editing results.}
	\label{fig:awm-straight-to-turn}
\end{figure}
	
\begin{figure}[!t]
	\centering
	\includegraphics[width=\hsize]{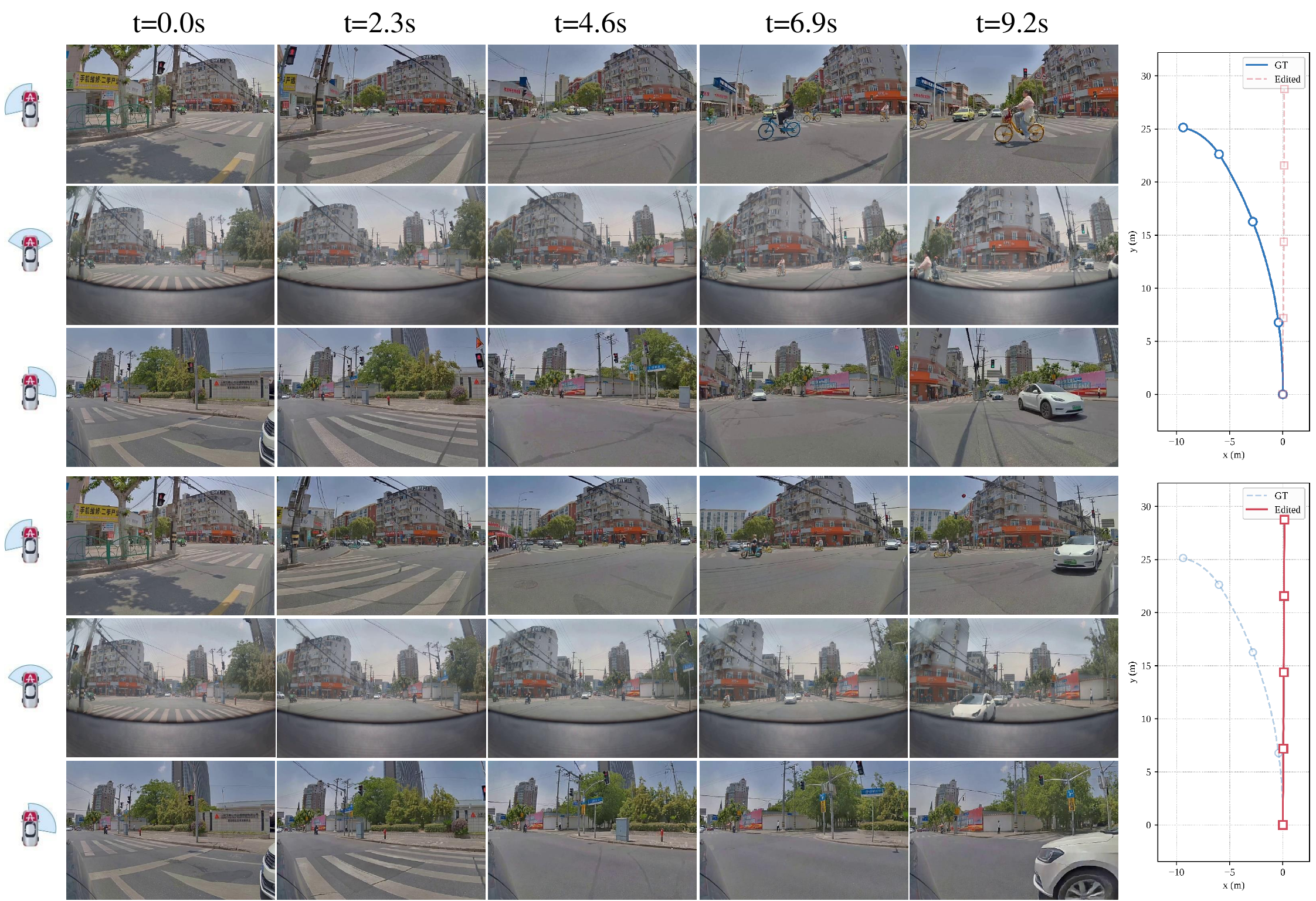}
	\caption{Turn-to-straight trajectory editing results.}
	\label{fig:awm-turn-to-straight}
\end{figure}

\begin{figure}[!t]
	\centering
	\includegraphics[width=\hsize]{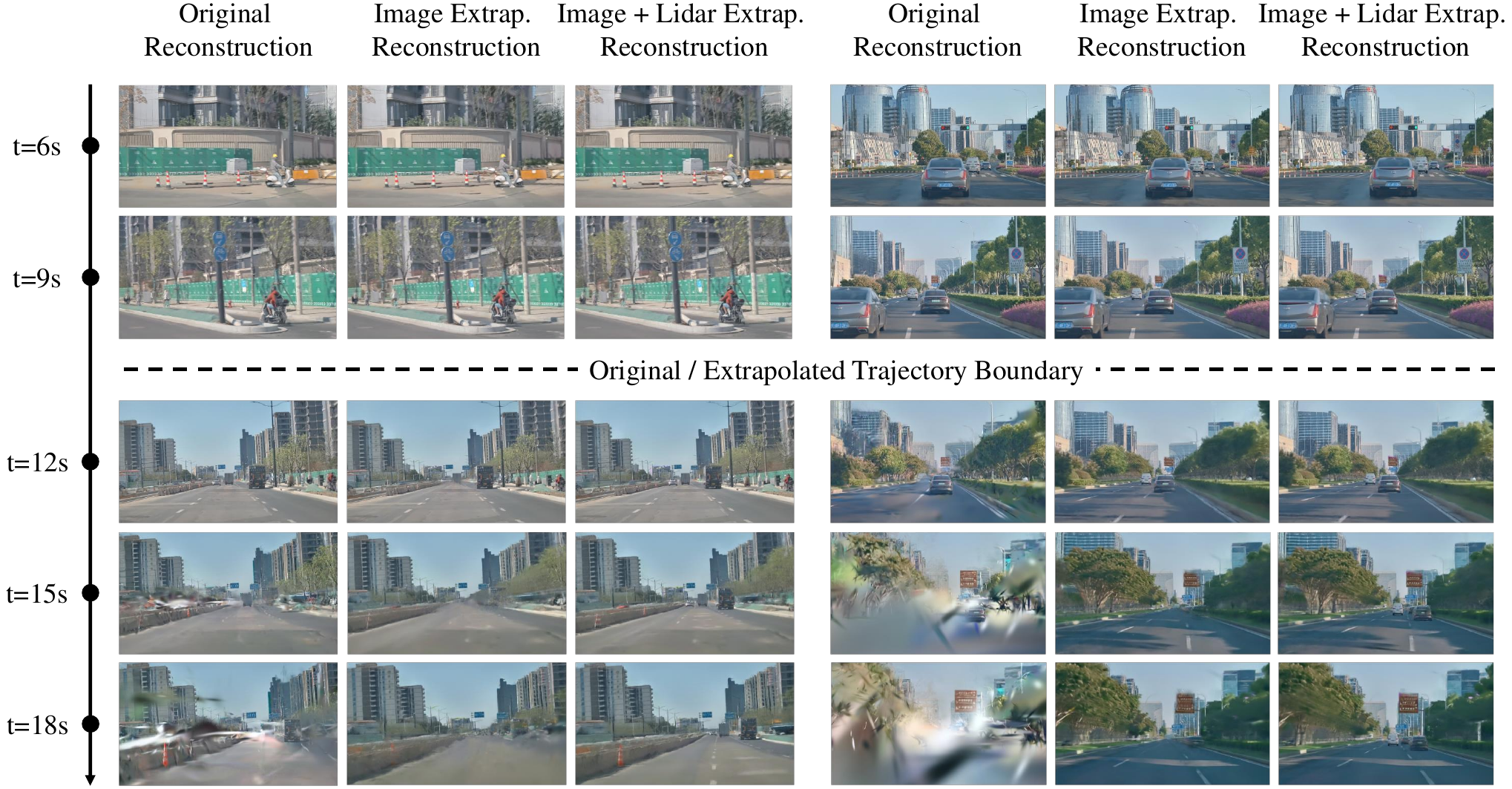}
	\caption{Reconstruction results on extrapolated trajectory.}
	\label{fig:3dgs}
\end{figure}

\textbf{Trajectory Editing.} We further evaluate the controllability of the model by modifying the input trajectories while retaining the original scene context. As shown in Fig.~\ref{fig:awm-lanechange} to~\ref{fig:awm-turn-to-straight}, we consider four representative edits: lane changing, overtaking, converting straight driving into a turn, and converting a turn into straight driving. For each case, we show the original sequence together with the generation conditioned on the edited trajectory. The generated vehicle motion follows the specified maneuver, while the road geometry and surrounding scene content remain consistent with the original scene. These results demonstrate that the model can generate plausible counterfactual observations beyond the trajectories recorded in the original data.

\subsubsection{3DGS Scene Extrapolation}
The preceding evaluation treats the action-conditioned world model as a standalone generative simulator. We now show that the same BehaviorFlow interface transfers to a reconstruction-based simulator without any change to its upstream output. Because BehaviorFlow generates structured trajectories rather than simulator-specific representations, the behavior-aware rollout that drives the generative world model can equally drive a 3D Gaussian Splatting pipeline. The multi-view videos and synchronized LiDAR rangemaps evaluated in Sec.~\ref{sec:exp_awm} are also the observations this pipeline consumes. Therefore, the generative and reconstruction branches form two ends of a single interface rather than two isolated systems. We use this property to address a persistent limitation of reconstruction-based simulators, namely their fixed spatial boundary.

We evaluate the proposed 3DGS scene extrapolation pipeline by extending each driving log beyond its final observed segment. BehaviorFlow first predicts the future traffic evolution, including trajectories for the ego vehicle and surrounding agents. The predicted traffic states are rendered into control videos that provide structured motion and behavior conditions for the action-conditioned world model. Based on these conditions, the world model generates multi-view camera observations and LiDAR point clouds along the extrapolated trajectory. These multimodal observations are appended to the original log before joint 3D Gaussian reconstruction.

To isolate the contributions of the generated image and LiDAR observations, we construct three reconstruction variants from the same driving sequence:
(1) Original Reconstruction, which uses only the original observed log;
(2) Image Extrapolation Reconstruction, which augments the original log with generated camera observations; and
(3) Image + LiDAR Extrapolation Reconstruction, which augments the original log with generated camera observations and LiDAR point clouds. Fig.~\ref{fig:3dgs} presents qualitative comparisons on two representative driving scenes. For each scene, the upper rows show viewpoints along the original observed trajectory. The lower rows show viewpoints along the extrapolated trajectory. 

Within the observed region, all variants achieve comparable reconstruction quality. This comparison indicates that the generated observations do not noticeably degrade reconstruction within the original coverage. Once the ego viewpoint moves beyond the observed region, the Original Reconstruction rapidly deteriorates. The degradation becomes more severe as the viewpoint progresses farther along the extrapolated trajectory. Typical failures include blurred textures, incomplete geometry, and prominent floater artifacts. These failures make the rendered observations unreliable for closed-loop simulation outside the original reconstruction boundary.

Both extrapolation-augmented variants maintain substantially higher visual quality in the extrapolated region. Image Extrapolation Reconstruction preserves the overall scene appearance while extending the usable rendering range. However, image-only supervision leaves weakly observed or occluded regions geometrically underconstrained. This limitation is particularly visible near dynamic objects, where incorrectly positioned Gaussian primitives may occlude foreground actors or interfere with downstream perception.

Adding generated LiDAR observations provides direct geometric constraints on the spatial positions of Gaussian primitives. As a result, the image-and-LiDAR variant produces more stable scene geometry, suppresses ground-level floaters, and improves the visibility of dynamic objects. These results demonstrate that generative image extrapolation is effective for extending the spatial coverage of 3DGS-based simulators, while the additional LiDAR supervision is crucial for maintaining geometric reliability in the extrapolated regions.

\begin{table*}[!t]
	\centering
	\caption{Comparison of results on the NAVSIM benchmark. The
		\textbf{best} and \underline{second-best} results are denoted by
		\textbf{bold} and \underline{underline}, respectively.}
	\vspace{-0.2cm}
	\label{tab:NAVSIM_main}
	\setlength{\tabcolsep}{4pt}
	\begin{adjustbox}{max width=\textwidth}
		\begin{tabular}{lHcccccc}
			\toprule
			\textbf{Methods} & \textbf{Venue} & \textbf{PDMS$\uparrow$} & \textbf{NC$\uparrow$} & \textbf{DAC$\uparrow$} & \textbf{EP$\uparrow$} & \textbf{TTC$\uparrow$} & \textbf{Comf.$\uparrow$} \\
			\midrule
			UniVLA  ~\citep{univla}    & RSS'25             & 81.7 & 96.9 & \textbf{91.1} & 76.8 & 91.7 & 96.7 \\
			AutoVLA ~\citep{zhou2026autovla}    & NeurIPS'25         & 89.1 & 98.4 & 95.6 & 81.9 & \textbf{98.0} & \underline{99.9} \\
			FSDrive  ~\citep{FSDrive}    & NeurIPS'25         & 85.1 & 98.2 & 93.8 & 80.1 & 93.3 & \underline{99.9} \\
			DriveVLA-W0 ~\citep{DriveVLA-W0} & ICLR'26            & 90.2 & 98.7 & \textbf{99.1} & 83.3 & 95.3 & 99.3 \\
			ReCogDrive ~\citep{li2025recogdrive}  & ICLR'26            & 90.8 & 97.9 & 97.3 & 87.3 & 94.9 & \textbf{100.0} \\
			SGDrive  ~\citep{li2026sgdrive}   & CVPR'26            & 91.1 & 98.6 & 97.8 & 85.8 & \underline{96.2} & \textbf{100.0} \\
			SpanVLA  ~\citep{zhou2026spanvla}    & arXiv'26  & 90.3 & \textbf{99.1} & 97.1 & 86.3 & 95.2 & \textbf{100.0} \\
			LatentVLA ~\citep{xie2026latentvla}& arXiv'26  & 92.4 & \underline{98.9} & 98.2& 88.2 & 96.0&\textbf{100.0} \\
			ChainFlow-VLA$^\dagger$$_{(train)}$~\citep{wang2026chainflow}   & arXiv'26 & 93.1 & {98.8} & 98.4 & \textbf{90.2} & 95.9 & \textbf{100.0} \\
			\midrule
			\textbf{Ours}$^\dagger$ & arXiv'26 & \textbf{93.3} & \underline{98.9} & \underline{98.7} & \underline{89.9} & \underline{96.2} & \textbf{100.0} \\
			\bottomrule
		\end{tabular}
	\end{adjustbox}
	\begin{flushleft}
		\footnotesize $^\dagger$ \textit{Note}: ChainFlow-VLA and our method are trained only with the first stage.
	\end{flushleft}
	\vspace{-0.5cm}
\end{table*}

\begin{table*}[!t]
	\centering
	\renewcommand{\tabcolsep}{4.75pt}
	\caption{Comparison on the NAVSIM benchmark. The \textbf{best} and \underline{second-best} results are denoted by
		\textbf{bold} and \underline{underline}, respectively.}
	\vspace{-0.3cm}
	\label{tab:recogdrive_navsim}
	\begin{tabular}{l|cccccc}
		\toprule
		\textbf{Methods}
		& \textbf{PDMS$\uparrow$}
		& \textbf{NC$\uparrow$}
		& \textbf{DAC$\uparrow$}
		& \textbf{EP$\uparrow$}
		& \textbf{TTC$\uparrow$}
		& \textbf{Comf.$\uparrow$} \\
		\midrule
		Constant Velocity
		& 20.6 & 68.0 & 57.8 & 19.4 & 50.0 & \textbf{100.0} \\
		Ego Status MLP
		& 65.6 & 93.0 & 77.3 & 62.8 & 83.6 & \textbf{100.0} \\
		\midrule
		VADv2-$\mathcal{V}_{8192}$~\citep{jiang2024vadv2}
		& 80.9 & 97.2 & 89.1 & 76.0 & 91.6 & \textbf{100.0} \\
		DrivingGPT~\citep{chen2025drivinggpt}
		& 82.4 & \textbf{98.9} & 90.7 & 79.7 & \textbf{94.9} & 95.6 \\
		UniAD~\citep{uniad}
		& 83.4 & 97.8 & 91.9 & 78.8 & 92.9 & \textbf{100.0} \\
		Transfuser~\citep{chitta2022transfuser}
		& 84.0 & 97.7 & 92.8 & 79.2 & 92.8 & \textbf{100.0} \\
		PARA-Drive~\citep{weng2024drive}
		& 84.0 & 97.9 & 92.4 & 79.3 & 93.0 & 99.8 \\
		DRAMA~\citep{yuan2024drama}
		& 85.5 & 98.0 & 93.1 & 80.1 & \underline{94.8} & \textbf{100.0} \\
		Hydra-MDP-$\mathcal{V}_{8192}$-W-EP~\citep{li2024hydra}
		& \underline{86.5} & 98.3 & \textbf{96.0} & 78.7 & 94.6 & \textbf{100.0} \\
		ReCogDrive$^{*}$~\citep{li2025recogdrive}
		& \underline{86.5} & 98.1 & 94.7 & \underline{80.9} & 94.2 & \textbf{100.0} \\
		\midrule
		\textbf{Ours}
		& \textbf{87.3} & \underline{98.4} & \underline{95.3} & \textbf{81.5} & \textbf{94.9} & \textbf{100.0} \\
		\bottomrule
	\end{tabular}
	\vspace{-0.3cm}
\end{table*}

\subsection{Action Model Results}
To assess the effectiveness of the data generated by BehaviorWorldGen, we select three representative action models and train each model on the 85K samples from the NAVSIM training split, augmented with 16K synthetic samples. Each model is trained using its corresponding configuration. Evaluation is conducted on NAVSIM~\citep{dauner2024navsim}, a large-scale autonomous driving benchmark that combines real-world driving data with a non-reactive simulation protocol for scalable evaluation. We report the official PDM Score (PDMS) and its component metrics, including no at-fault collision (NC), drivable area compliance (DAC), ego progress (EP), time to collision (TTC), and driving comfort (Comf.). The selected models cover two representative planning paradigms: ChainFlow-VLA~\citep{wang2026chainflow} and ReCogDrive~\citep{li2025recogdrive} are vision-language-action planning models, whereas DiffusionDrive~\citep{liao2025diffusiondrive} is an end-to-end driving model.

\subsubsection{Vision-Language-Action Model}

\textbf{ChainFlow-VLA.} We evaluate BehaviorWorldGen on ChainFlow-VLA under its first-stage training setting. As shown in Tab.~\ref{tab:NAVSIM_main}, our method achieves the highest PDMS among compared VLA approaches. After training with the behavior-aware data generated by BehaviorWorldGen, ChainFlow-VLA performs better on the safety-related NC, DAC, and TTC metrics. More importantly, the augmented model produces fewer catastrophic zero-score outcomes caused by collisions or severe violations of the drivable area. Such failures have a disproportionate effect on PDMS because they can invalidate a reasonable trajectory. Therefore, BehaviorWorldGen shifts ChainFlow-VLA toward safer interaction-aware decisions while preserving its strong planning capability.


\textbf{ReCogDrive.} 
We further apply BehaviorWorldGen to ReCogDrive~\citep{li2025recogdrive} to validate whether the improvement transfers across VLA architectures. Our reproduction omits the reinforcement-learning stage of the original method. As shown in Tab.~\ref{tab:recogdrive_navsim}, its PDMS increases to 87.3, yielding an absolute improvement of 0.8 points after training with the behavior-aware data generated by BehaviorWorldGen. The component metrics improve consistently. TTC exhibits the largest gain of 0.7 points, followed by DAC and EP with 0.6 points each, then NC with 0.3 points. The gains in NC and TTC indicate fewer collision-related failures with safer temporal margins, whereas the DAC gain reflects stronger drivable-area compliance. The comfort score remains unchanged, so the improvement does not compromise trajectory smoothness. Therefore, the advantage of BehaviorWorldGen is not tied to a specific VLA architecture.

\subsubsection{End-to-End Planner}
\textbf{Overall performance.}
As reported in Tab.~\ref{tab:main_navsim}, our reproduced DiffusionDrive baseline achieves a PDMS of 87.7. After training with the behavior-aware data generated by BehaviorWorldGen, its PDMS increases to 88.6, yielding an absolute improvement of 0.9 points. Consistent gains are observed across all component metrics. Among the component metrics, EP exhibits the largest improvement of 0.7 points, followed by DAC and TTC with gains of 0.6 points each. The gains in NC, DAC, and TTC indicate improved safety-related performance with stronger drivable-area compliance. Meanwhile, the improvement in EP suggests that the policy does not become overly conservative. The unchanged comfort score further shows that the overall gain does not compromise trajectory smoothness.

\begin{table*}[!t]
	\centering
	\renewcommand{\tabcolsep}{4.75pt}
	\caption{Comparison on the NAVSIM benchmark. The
		\textbf{best} and \underline{second-best} results are denoted by
		\textbf{bold} and \underline{underline}, respectively.}
	\vspace{-0.3cm}
	\label{tab:main_navsim}
	\begin{tabular}{l|cccccc}
		\toprule
		\textbf{Methods}
		& \textbf{PDMS$\uparrow$}
		& \textbf{NC$\uparrow$}
		& \textbf{DAC$\uparrow$}
		& \textbf{EP$\uparrow$}
		& \textbf{TTC$\uparrow$}
		& \textbf{Comf.$\uparrow$} \\
		\midrule
		
		UniAD~\citep{uniad}
		& 83.4 & 97.8 & 91.9 & 78.8 & 92.9 & \textbf{100.0} \\
		PARA-Drive~\citep{weng2024drive}
		& 84.0 & 97.9 & 92.4 & 79.3 & 93.0 & 99.8 \\
		LTF~\citep{chitta2022transfuser}
		& 83.8 & 97.4 & 92.8 & 79.0 & 92.4 & \textbf{100.0} \\
		Transfuser~\citep{chitta2022transfuser}
		& 84.0 & 97.7 & 92.8 & 79.2 & 92.8 & \textbf{100.0} \\
		DRAMA~\citep{yuan2024drama}
		& 85.5 & 98.0 & 93.1 & 80.1 & \textbf{94.8} & \textbf{100.0} \\
		VADv2-$\mathcal{V}_{8192}$~\citep{jiang2024vadv2}
		& 80.9 & 97.2 & 89.1 & 76.0 & 91.6 & \textbf{100.0} \\
		Hydra-MDP-$\mathcal{V}_{8192}$~\citep{li2024hydra}
		& 83.0 & 97.9 & 91.7 & 77.6 & 92.9 & \textbf{100.0} \\
		Hydra-MDP-$\mathcal{V}_{8192}$-W-EP~\citep{li2024hydra}
		& 86.5 & \textbf{98.3} & 96.0 & 78.7 & \underline{94.6} & \textbf{100.0} \\
		DiffusionDrive$^*$~\citep{liao2025diffusiondrive}
		& \underline{87.7} & \underline{98.1} & \underline{96.1}
		& \underline{82.1} & 94.2 & \textbf{100.0} \\
		\midrule
		\textbf{Ours}
		& \textbf{88.6} & \textbf{98.3} & \textbf{96.7}
		& \textbf{82.8} & \textbf{94.8} & \textbf{100.0} \\
		\bottomrule
	\end{tabular}
\end{table*}

\begin{table}[t]
	\centering
	\caption{Comparison with DiffusionDrive on low-PDMS scenarios.}
	\vspace{-0.3cm}
	\label{tab:low_pdms_comparison}
	\begin{tabular}{l|l|cccccc}
		\toprule
		\textbf{Methods}
		& \textbf{Range}
		& \textbf{PDMS$\uparrow$}
		& \textbf{NC$\uparrow$}
		& \textbf{DAC$\uparrow$}
		& \textbf{EP$\uparrow$}
		& \textbf{TTC$\uparrow$}
		& \textbf{Comf.$\uparrow$} \\
		\midrule
		
		\multirow{4}{*}{DiffusionDrive}
		& $[0, 0.15)$
		& 0.0 & 65.5 & 32.0 & 0.1 & 60.7 & \textbf{100.0} \\
		\cmidrule(lr){2-8}
		
		& $[0.15, 0.3)$
		& 20.8 & 78.7 & 54.7 & 33.6 & 51.6 & \textbf{100.0} \\
		\cmidrule(lr){2-8}
		
		& $[0.3, 0.45)$
		& 38.2 & 90.0 & 68.7 & 48.1 & 46.0 & \textbf{100.0} \\
		
		\midrule
		
		\multirow{4}{*}{\textbf{Ours}}
		& $[0, 0.15)$
		& \textbf{34.8}
		& \textbf{78.0} & \textbf{59.9} & \textbf{34.8}
		& \textbf{68.6} 
		& \textbf{100.0} \\
		\cmidrule(lr){2-8}
		
		& $[0.15, 0.3)$
		& \textbf{39.4}
		& \textbf{89.5} & \textbf{84.2} & \textbf{66.3}
		& \textbf{68.4} 
		& \textbf{100.0} \\
		\cmidrule(lr){2-8}
		
		& $[0.3, 0.45)$
		& \textbf{60.0}
		& \textbf{96.7}& \textbf{80.0} & \textbf{65.7}
		& \textbf{73.3} 
		& \textbf{100.0} \\
		
		\bottomrule
	\end{tabular}%
	\vspace{-0.3cm}
\end{table}

\textbf{Performance on low-score scenarios.} Aggregate NAVSIM results are dominated by routine driving cases, which can obscure the effect of targeted behavior generation. We consequently evaluate the augmented model on scenarios where the normalized PDMS of the original DiffusionDrive is below 0.45. As shown in Tab.~\ref{tab:low_pdms_comparison}, BehaviorWorldGen improves planning performance across all three intervals. The most pronounced recovery occurs in the $[0,0.15)$ interval. The original DiffusionDrive nearly collapses to a zero PDMS, indicating a catastrophic failure regime. After training with behavior-aware data, the model recovers meaningful planning performance on the same scenarios. This result shows that BehaviorWorldGen can correct severe failures rather than merely refine already successful predictions.

The component metrics reveal different sources of improvement across the three difficulty levels. In the two more difficult intervals, the strongest gains appear in DAC and EP. This suggests that the augmented policy avoids severe drivable-area violations while recovering effective route progress. In the $[0.3,0.45)$ interval, the most prominent improvement occurs in TTC. This result indicates safer temporal margins in scenarios where the baseline already makes partial progress. NC also improves consistently across all intervals, reflecting fewer collision-related failures. The comfort score remains unchanged throughout the comparison. Therefore, the improvements in safety and progress are not achieved through less smooth trajectories. Overall, the substantially larger gains on low-PDMS scenarios confirm that BehaviorWorldGen concentrates its benefit on difficult failure cases that are underrepresented by aggregate benchmark results.

\begin{figure}[!t]
    \vspace{-0.4cm}
	\centering
	\includegraphics[width=\hsize]{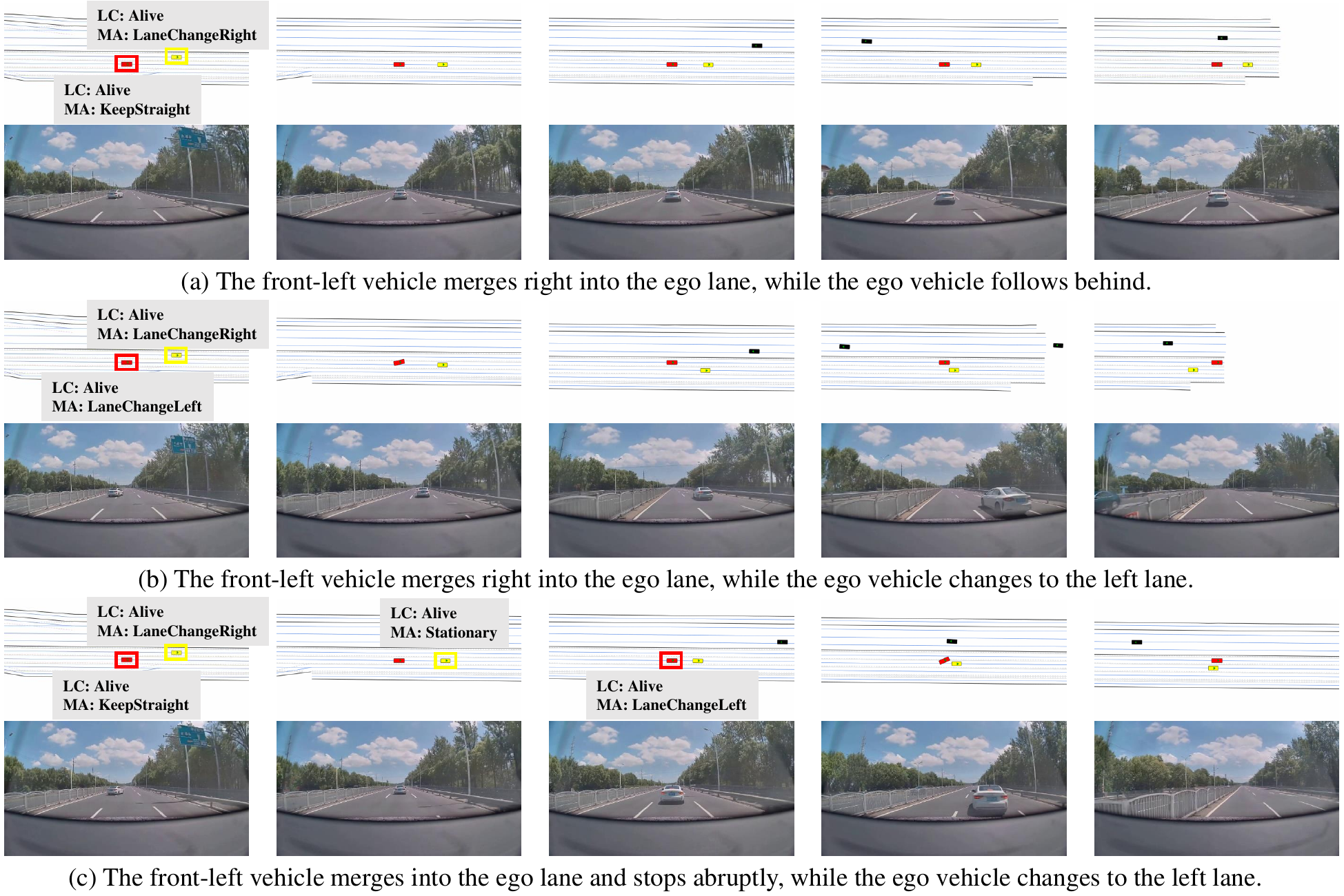}
	\caption{Qualitative results on a straight-driving scene of BehaviorFlow under  behavior editing.}
	\label{fig:behaviorFlow_editing_scene1}
\end{figure}

\begin{figure}[!t]
	\centering
	\includegraphics[width=\hsize]{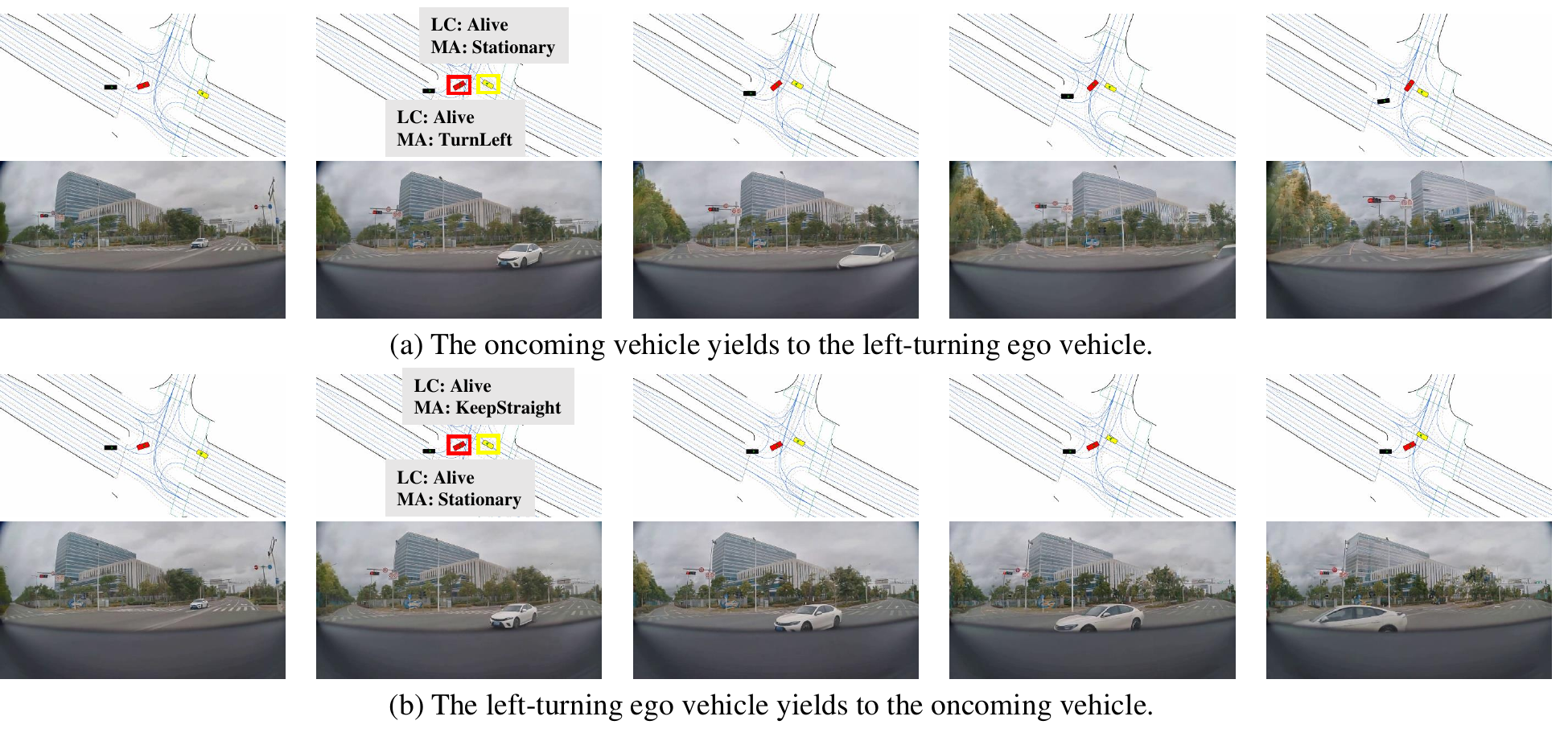}
	\caption{Qualitative results on a left-turn scene of BehaviorFlow under  behavior editing.}
	\label{fig:behaviorFlow_editing_scene2}
\end{figure}

\subsection{BehaviorFlow Results}\label{sec:behaviorflow_results}
We qualitatively evaluate the controllability and interaction consistency of BehaviorFlow on our private dataset using the traffic rollouts shown in Fig.~\ref{fig:behaviorFlow_editing_scene1} and Fig.~\ref{fig:behaviorFlow_editing_scene2}. These figures present two representative scenarios: a straight-driving scenario with a cut-in vehicle and an intersection scenario involving an ego left turn. For each scenario, different meta-action constraints are applied to the interacting agents. BehaviorFlow then regenerates their future states under the specified interventions.

In the straight-driving scenario, a vehicle initially travels ahead and to the left of the ego vehicle. We construct three variants with different behaviors following its right lane change into the ego lane. In the first variant, the front-left vehicle merges into the ego lane and continues at a constant speed. The ego vehicle follows behind it. In the second variant, the front-left vehicle performs the same lane change and maintains a constant speed. The ego vehicle instead changes to the left lane. In the third variant, the front-left vehicle merges into the ego lane, and comes to an emergency stop. The ego vehicle responds by changing to the left lane and driving around the stopped vehicle. These variants demonstrate that BehaviorFlow can generate distinct following and emergency-avoidance behaviors from the same initial scene.

The intersection scenario evaluates controllable yielding between the left-turning ego vehicle and an oncoming vehicle traveling straight. We generate two variants with opposite passing orders. In both variants, the two vehicles decelerate before reaching the potential collision point. In the first variant, the oncoming vehicle yields, allowing the ego vehicle to complete its left turn first. The oncoming vehicle then proceeds straight through the intersection. In the second variant, the ego vehicle yields while the oncoming vehicle crosses the intersection first. The ego vehicle subsequently completes its left turn. Despite sharing the same initial scene and intended routes, the two variants produce different yielding relations while avoiding collisions.

Across both scenarios, the generated agents realize the specified maneuvers while adapting coherently to one another.  These results demonstrate that BehaviorFlow can generate diverse and interaction-consistent rollouts from the same scene context, providing controllable training scenarios for world simulation and subsequent action-model refinement.

\section{Conclusion}
We introduced BehaviorWorldGen, a framework that closes the loop between action models and world simulators through controllable behavior-aware structured world generation. At its core, BehaviorFlow provides meta-action-conditioned control over agent lifecycles and high-level behaviors to generate interaction-consistent multi-agent rollouts. This design addresses a fundamental limitation of existing world simulators, which mainly condition on ego actions and cannot model responsive surrounding-agent behaviors. Their rollouts consequently provide distorted interaction feedback with limited coverage of safety-critical long-tail scenarios. BehaviorWorldGen uses structured trajectories as the interface between action models and world simulators. The resulting rollouts are rendered into realistic observations, then paired with corrected interaction-aware trajectories for policy refinement. This structured interface makes the framework compatible with different policy architectures and rendering mechanisms. Experiments on world generation, 3DGS scene extrapolation, and end-to-end planning demonstrate the effectiveness of controllable behavior generation for action-model improvement. Multimodal image and LiDAR extrapolation further extends the reliable operating range of reconstruction-based simulators.

\hypertarget{author-list}{}
\section*{Author List}
\textbf{World Simulator:}
\begin{indented}
  Generation World Model: Zhuo Zhang $\cdot$ Jiaqi Wang $\cdot$ Haowen Cui $\cdot$ Chuanye Wang $\cdot$ Zhongyang Zhu $\cdot$  Yulong Zheng 
  
  Reconstruction World Model: Haining Guan 
\end{indented}
\textbf{Action Model:} Tingguang Zhou $\cdot$ Zhen Yang

\textbf{BehaviorFlow:} Jiaqi Wang $\cdot$ Xuefeng Chen  $\cdot$ Haining Guan  $\cdot$ Tianchen Deng

\textbf{Project Advisor:} Lixia Shen $\cdot$ Bo Dai $\cdot$ Hangning Zhou $\cdot$ Feiyang Tan

\textbf{Project Leader:} Jiajun Zhu $\cdot$ Xiyang Wang $\cdot$ Xiwu Chen

\section*{Acknowledgements}
We sincerely thank the following individuals for their valuable contributions and support:  Wei Tan $\cdot$ Chao Lu $\cdot$ Xi Su $\cdot$ Zihan Lin $\cdot$ Qun Wu $\cdot$  John Shao  $\cdot$ Wenqin Liu $\cdot$ Xinjing He  $\cdot$ Wenpan Li $\cdot$ Yao He $\cdot$ Yang Hu $\cdot$ BaoCheng Geng $\cdot$ Yuanlin Jin $\cdot$ Jianpeng Gan

\bibliography{behaviorworldgen}
\bibliographystyle{behaviorworldgen}

\newpage
\appendix
\begin{center}
{\Large\bfseries Appendix}
\end{center}
\section{3D Gaussian Scene Reconstruction}\label{app:3dgs}
The reconstruction system builds upon the multi-representation scene decomposition of OmniRe~\citep{chen2025omnire}, where static environments and dynamic actors are modeled by separate Gaussian representations and rendered in a unified scene graph. Three main adaptations are introduced for closed-loop simulation: RoGS-based road modeling, object-level reconstruction and refinement of dynamic vehicles, and MVSA-based geometric completion for regions beyond the coverage of forward-facing LiDAR.
\subsection{Road-Surface Reconstruction}

Conventional 3DGS optimizes Gaussian positions mainly through photometric supervision. Although this can reproduce sharp road lane lines from observed viewpoints, road primitives may drift from the physical surface to reduce image-space errors. The resulting geometric inaccuracies become pronounced under lateral viewpoint shifts, causing road lane lines to blur or distort—a critical failure for closed-loop simulation when the ego vehicle deviates from the logged trajectory.
\begin{figure}[!h]
	\centering
	\includegraphics[width=1\hsize]{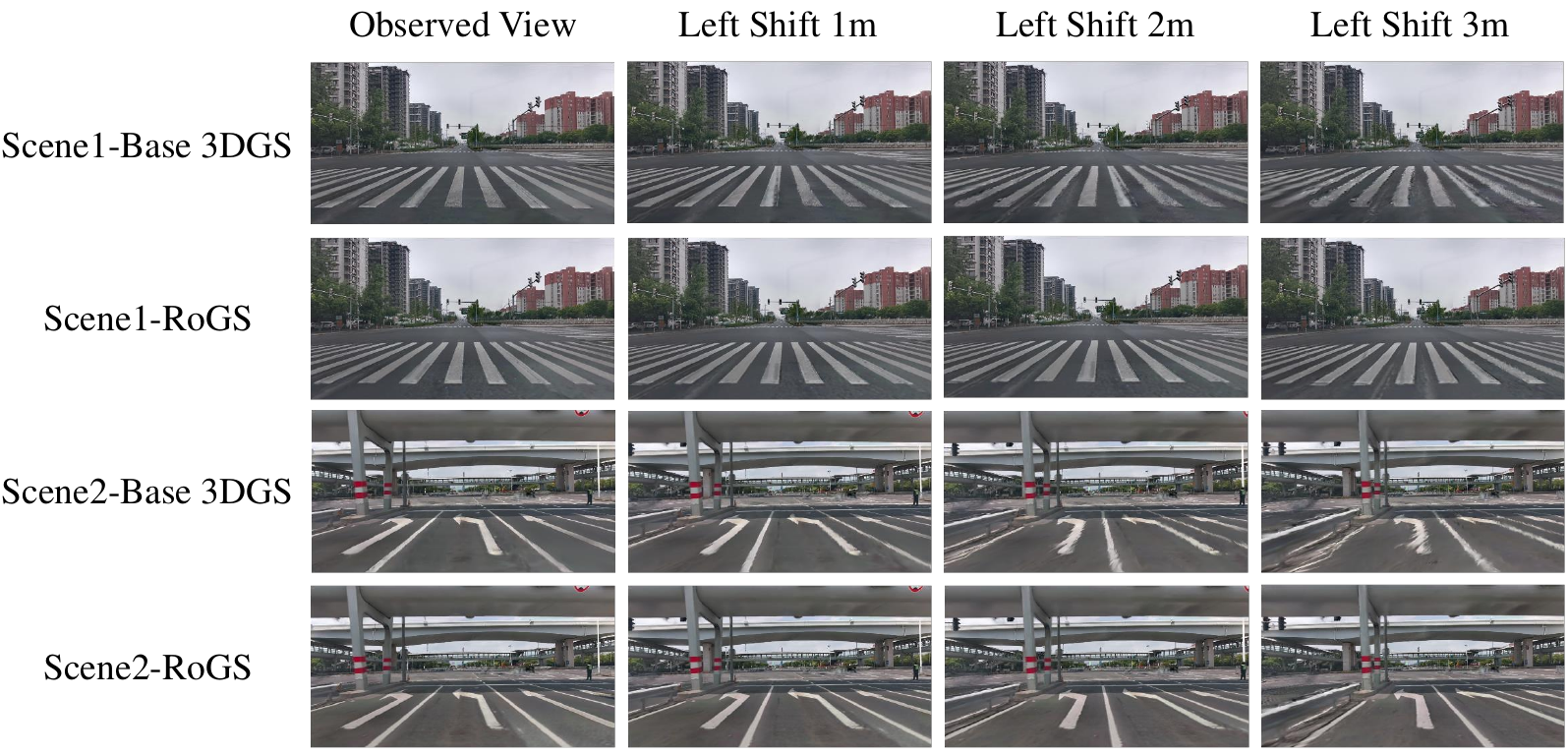}
	\caption{Road reconstruction under lateral viewpoint shifts.}
	\label{road_rogs}
\end{figure}

RoGS instead represents the road using dense 2D Gaussian primitives on a regular bird's-eye-view grid without relying on LiDAR. The ego trajectory recovered from camera poses and extrinsics determines the grid extent, and the nearest ego pose initializes the height and orientation of each primitive. Other Gaussian attributes are learned from image observations. During optimization, horizontal positions remain fixed, while heights are optimized with local smoothness regularization. This structured height-field representation prevents arbitrary horizontal drift, preserves continuous road geometry, and improves the rendering stability of road lane lines under lateral novel views. As shown in Fig. \ref{road_rogs}, the road reconstruction preserves sharp and geometrically consistent lane markings under lateral viewpoint shifts, whereas the original 3DGS exhibits increasing distortion away from the observed trajectory. 

\subsection{Dynamic-Vehicle Reconstruction and Refinement}

\begin{figure}[!h]
	\centering
	\includegraphics[width=1\hsize]{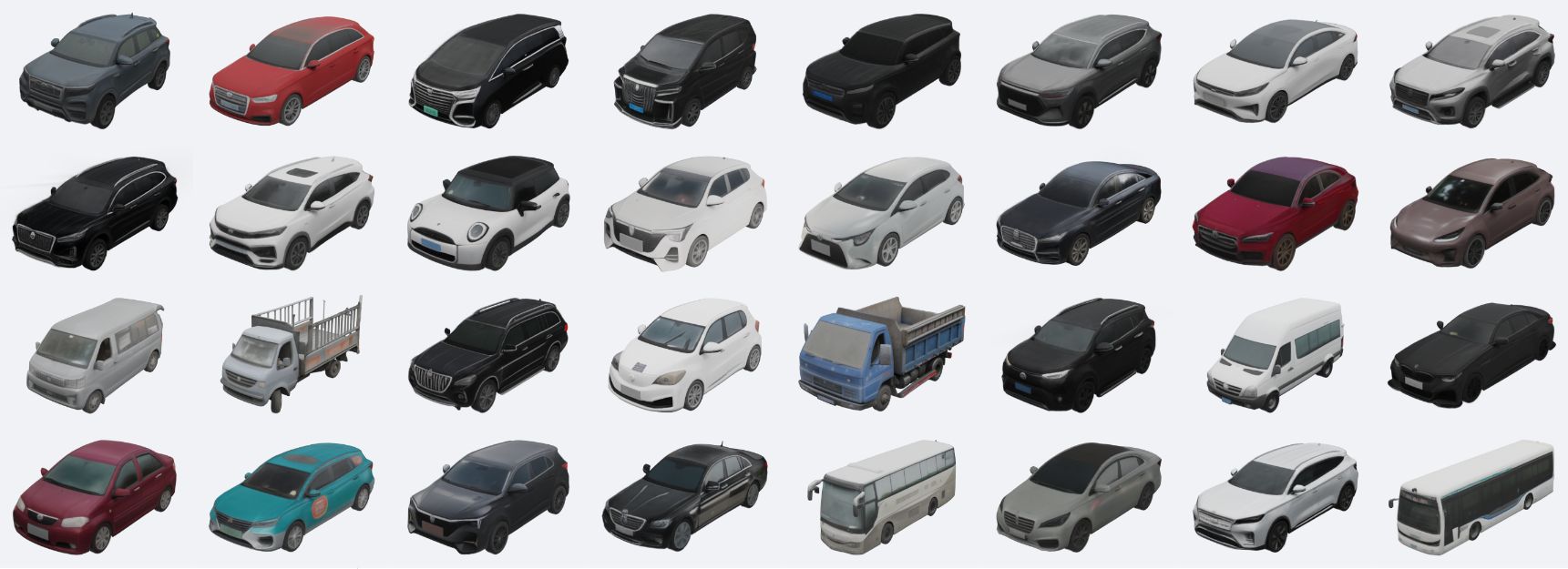}
	\caption{Representative object-centric vehicle reconstructions. The reconstructed assets cover diverse vehicle categories, shapes, and appearances.}
	\label{object_recon}
\end{figure}

Direct scene-level optimization often produces incomplete vehicles because each actor is observed from limited viewpoints and is frequently occluded. Therefore, MVSAM3D reconstructs each valid vehicle as an independent object-centric 3D Gaussian model. For each track, high-quality observations are selected across cameras and timestamps according to visibility, projected object size, segmentation quality, and viewpoint diversity. The selected RGB crops and masks are jointly processed to recover more complete vehicle geometry and appearance. Fig. \ref{object_recon} shows representative object-centric reconstructions across diverse vehicle categories and appearances, demonstrating the coverage of the resulting vehicle asset collection.

\begin{figure}[!t]
	\centering
	\includegraphics[width=1\hsize]{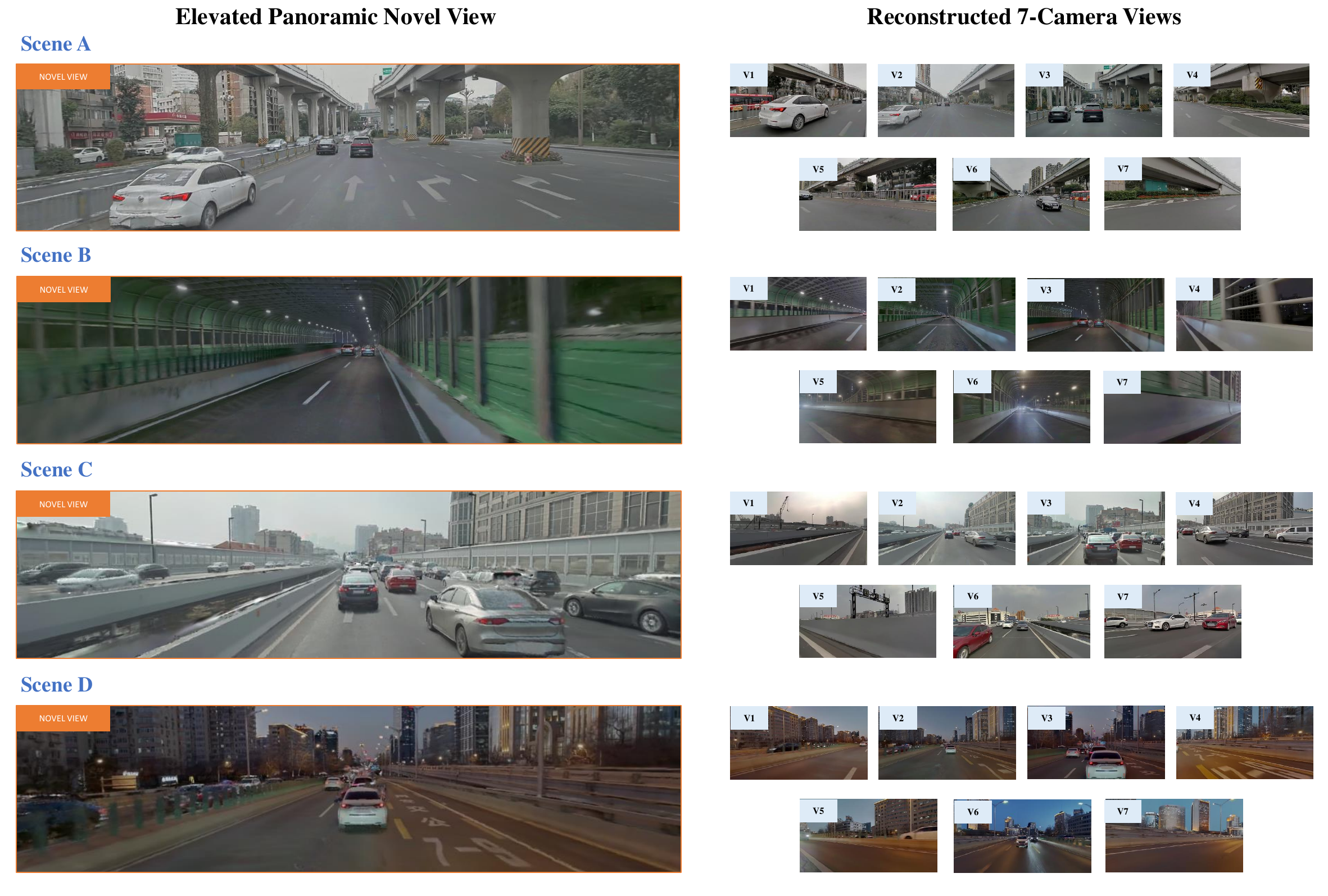}
	\caption{Overall scene reconstruction results. Elevated panoramic novel views and reconstructed seven-camera views are shown for diverse driving environments.}
	\label{recon_main}
\end{figure}

Because the reconstructed object lacks a reliable metric scale, its canonical orientation and dimensions are first aligned with the average 3D bounding-box size of the corresponding track. The object is then inserted into the scene with its original track ID and motion trajectory.

After the base scene converges, the remaining scene components are fixed and the inserted vehicles follow a three-stage coarse-to-fine refinement schedule. Low-frequency appearance, object pose, and residual three-axis scale are first optimized for coarse photometric and geometric alignment. The pose and scale are then fixed while higher-order spherical harmonics and opacity recover view-dependent appearance details. Finally, the vehicle representation is fixed and only its ground-contact shadow is refined. This schedule separates object alignment, appearance enhancement, and scene integration while retaining the complete structure recovered by MVSAM3D.

\subsection{Rear-View Geometry Completion with MVSA}

Reliable Gaussian initialization is critical to high-quality 3DGS reconstruction. For platforms equipped with predominantly forward-facing solid-state LiDAR, the rear-view region receives no LiDAR observation at the beginning of a sequence. The resulting lack of geometric initialization severely degrades rear-view reconstruction. An MVSA-based visual depth completion strategy is introduced to provide dense geometric initialization and supervision for this region.

MVSA predicts dense depth from calibrated multi-camera observations without relying on LiDAR. Since raw predictions can be unreliable in occluded or weakly textured regions, a pose-aware multi-view geometric consistency filter is further designed. Each reference-view depth is back-projected into 3D, projected into neighboring source views, and compared with the corresponding source-view depth predictions. Estimates without sufficient cross-view agreement are discarded before reconstruction.

The filtered depths serve two complementary purposes. They are back-projected to initialize background Gaussian primitives in the rear-view region, compensating for the missing LiDAR points, and are also used as dense depth supervision to constrain Gaussian positions during reconstruction. Together, these two uses resolve the geometric cold-start problem in the initial rear-view region.

\subsection{Camera Appearance and Environment Modeling}

A learnable bilateral grid compensates for camera-dependent ISP and exposure differences, improving cross-camera appearance consistency. The distant sky is represented using a learnable cubemap and composited with the Gaussian scene during rendering.

\subsection{Summary}
Together, the structured road representation, object-centric vehicle reconstruction, and MVSA-based rear-view completion combined with camera-aware appearance correction and sky modeling improve geometric consistency, scene completeness, and rendering quality across observed and novel viewpoints. 

Fig. \ref{recon_main} presents representative results of the overall reconstruction across diverse driving environments, including urban roads, highways, dense traffic, and nighttime scenes. The resulting system enables high-fidelity and geometrically complete driving-scene reconstruction, providing a reliable foundation for closed-loop simulation.

\end{document}